\documentclass[letterpaper, 10 pt, conference]{template/ieeeconf}  %

\IEEEoverridecommandlockouts                              %

\usepackage{amsmath} %
\usepackage{amssymb}  %
\usepackage{graphicx}
\usepackage{hyphenat}
\usepackage[table]{xcolor}
\usepackage[font=small,belowskip=0pt, aboveskip=6pt]{caption}
\usepackage{subcaption}

\usepackage{multirow}
\usepackage{booktabs,amsfonts,dcolumn}
\usepackage{tabularx}
\usepackage{makecell}
\usepackage{hyperref}
\usepackage{comment}
\usepackage{diagbox}
\usepackage{multirow}
\usepackage{calrsfs}

\usepackage{xspace}
\usepackage[group-separator={,},group-minimum-digits=4]{siunitx}

\DeclareMathAlphabet{\pazocal}{OMS}{zplm}{m}{n}

\long\def\invis#1{}

\newcommand\fig[1]{Fig. \ref{#1}}

\makeatletter
\DeclareRobustCommand\onedot{\futurelet\@let@token\@onedot}
\def\@onedot{\ifx\@let@token.\else.\null\fi\xspace}

\def\etal{\emph{et al}\onedot}
\makeatother

\title{\LARGE \bf
Mapping Pamir: Multi-Session Visual-Inertial SLAM and 3D Reconstruction of an Underwater Shipwreck
}

\author{Michalis Chatzispyrou$^{a*}$, Luke Horgan$^{b*}$, Hyunkil Hwang$^{a*}$, Harish Sathishchandra$^{b*}$, Chinmay Burgul$^{a}$,	 \\
Monika Roznere$^c$, Alberto Quattrini Li$^d$, Philippos Mordohai$^b$, Ioannis Rekleitis$^a$%
\thanks{$^*$The first four authors have contributed equally to this work and are listed in alphabetical order. \enspace
$^a$University of Delaware, Newark, DE, USA, {\tt\footnotesize \{michalis,hkhwang,cmburgul,yiannisr\}@udel.edu.} \enspace
$^b$Stevens Institute of Technology, Hoboken, NJ, USA, {\tt\footnotesize \{lhorgan,hsathish,pmordoha\}@stevens.edu} \enspace
$^c$Binghamton University, Binghamton, NY, USA {\tt\footnotesize mrozner1@binghamton.edu} \enspace
$^d$Dartmouth College, Hanover, NH, USA {\tt\footnotesize alberto.quattrini.li@dartmouth.edu}}%
\thanks{This research has been supported in part by the National Science Foundation under grants 1943205, 2024541, 2024653, and 2024741. The authors are also grateful for equipment support by Halcyon Dive Systems, Teledyne FLIR LLC, and KELDAN GmbH lights.}%
}

\begin{document}

\maketitle
\begin{abstract}
This paper presents a framework for multi-session mapping of underwater environments utilizing an affordable action camera. The Visual-Inertial data are augmented by water depth recordings from a dive computer. SVIn2, an open-source VI-SLAM framework is utilized to generate a trajectory and a sparse reconstruction for each session. Utilizing the keyframes extracted from SVIn2, and the estimated camera poses, a Structure-from-Motion (SfM) framework -- COLMAP -- is employed for global optimization and produce a dense reconstruction of the target environment. The presence of calibration targets at fixed locations, when available, is used to estimate the coordinate transformation between different data collection sessions, thus transforming the different sessions into the same coordinate frame. The proposed pipeline is employed for the mapping of a shipwreck off the coast of Barbados. For the first time, both the exterior and the accessible interior parts of the wreck were mapped in two sessions, while a third session employed two cameras with different fields of view.  
\end{abstract}

\section{Introduction}
Accurate mapping of underwater structures is crucial for several domains, including underwater archaeology~\cite{eriksson2017mars,mahon2011reconstructing,demesticha20144th,eustice2006visually}, off-shore energy platform inspection~\cite{gillies2015close}, and environmental monitoring~\cite{david2021structure}. However, underwater vision has proven to be extremely challenging~\cite{QuattriniLiISERVO2016,JoshiIROS2019,islam2024computer}. Additionally, the deployment of an autonomous underwater vehicle (AUV) is expensive and time consuming. Early work proposed the use of an inexpensive action camera~\cite{JoshiICRA2022} to deploy SVIn2~\cite{RahmanIJRR2022}, a robust Visual-Inertial (VI) SLAM framework based on OKVIS~\cite{okvis}. The advantage of the specific camera (GoPro Hero9-Hero13) is that it encodes video at 30 fps (or higher) and Inertial Measurement Unit (IMU) data at 100 Hz in a single video file, thereby allowing VIO and VI-SLAM algorithms to be run on it. Please refer to the work of Joshi~\etal~\cite{JoshiICRA2022} for more details and a comparison of different open-source VIO/VI-SLAM packages. From the formulation of the VI-SLAM problem, roll and pitch are observable, but the position and yaw orientation are not~\cite{hesch2014camera,yang2019observability}. Although in SVIn2~\cite{RahmanIJRR2022} a water-depth sensor was used, the GoPro action camera does not have one. However, since most divers carry a dive computer that records the depth in the water during the dive, this information is available for each deployment.  

The challenging nature of underwater environments makes extensive deployments quite difficult; nitrogen loading and the danger of decompression sickness in conjunction with the amount of breathing gas a diver can carry limit the available time underwater. This requires breaking the data collection over multiple dives. Most VI-SLAM approaches generate a sparse reconstruction of the environment, while dense reconstructions often require a synchronized stereo camera setup~\cite{WangICRA2023}. On the other hand, Structure-from-Motion (SfM) frameworks that perform global optimization, such as COLMAP~\cite{colmap_sfm}, are extremely slow for large numbers of images. 
Utilizing all the images of a video sequence, at over 100k frames per hour, is prohibitively expensive with COLMAP, while uniform sampling of the video results in tracking difficulties due to fast motions etc.
In this work, we propose to utilize the \textit{keyframes} generated by SVIn2 as input for bundle adjustment in COLMAP. Nearby keyframes have adequate overlap between them while maintaining sufficient spatial separation to provide large baselines for triangulation. Furthermore, the trajectory produced by SVIn2 is used as a pose prior in COLMAP, thus providing correct scale for the resulting reconstruction and injecting the IMU data into COLMAP. The Z-axis (depth) estimates from SVIn2 are corrected using the water depth measurements from a dive computer, in order to have access to the absolute values of the z-axis. In the case of multiple sessions where common fiducial markers (targets) remain in fixed locations, the coordinate transformations from the camera to the targets are utilized to bring the different trajectories into a common frame of reference. 

\begin{figure}[t]
     \centering
     {\includegraphics[width=0.9\columnwidth]{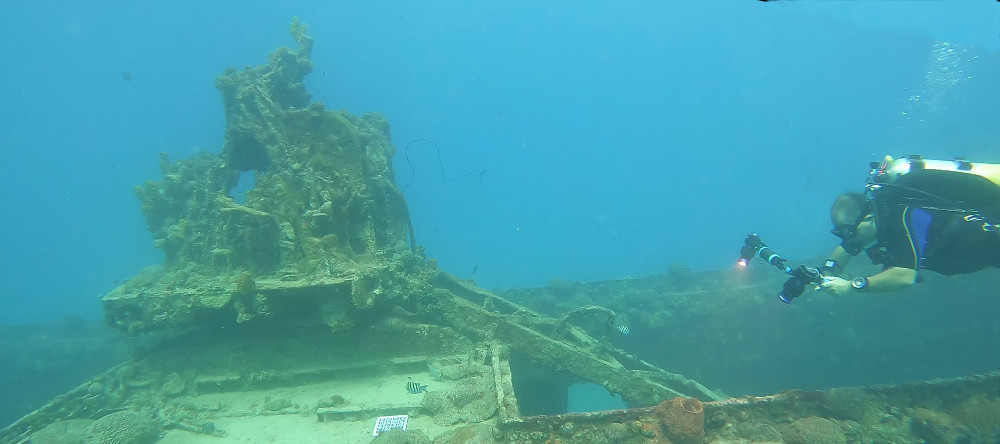}}
     \caption{GoPro setup deployed over the Pamir shipwreck, Barbados.}
     \label{fig:beauty}
     \vspace{-2em}
 \end{figure}

We apply the proposed framework to mapping the Pamir shipwreck off the coast of Barbados. From more than two hours (167 minutes) of GoPro video (\num{300000}+ frames)  captured over three separate dives, our results demonstrate a successful mapping of the wreck's exterior and interior parts. 
 
This paper makes the following contributions: a)~Introducing accurate z-axis absolute measurements from a dive computer into the VIO/VI-SLAM process. b)~Introducing a new methodology for frame selection to be used by COLMAP~\cite{colmap_sfm}, eliminating discontinuities (multiple sub-models) produced by rapid field of view changes. c)~Introducing correct scale to monocular bundle adjustment by utilizing the SVIn2 produced camera poses. Additionally, four VIO datasets (including water depth) from a shipwreck are released to the community.

\section{Related Work}\label{sec:related}

In this section, we review prior research in Structure-from-Motion (SfM), Simultaneous Localization and Mapping (SLAM), dense 3D reconstruction, and multi-session mapping  in underwater environments. We distinguish between \textbf{sparse} and \textbf{dense} approaches. The former estimate camera parameters (if a calibration process has not been previously carried out), 6D camera poses, and a 3D point cloud (by reconstructing image features tracked/detected across multiple images). Dense methods receive camera poses, and sometimes the sparse point cloud, and compute depth for almost every pixel in the images. We also review pipelines covering both the sparse and dense aspects of 3D reconstruction.

State estimation underwater is challenging due to color saturation, floating particulates, and limited visibility \cite{JoshiIROS2019}. 
Beall \etal. \cite{Beall} demonstrated accurate sparse 3D reconstruction of underwater structures, such as corals, from synchronized  high definition videos collected using a wide baseline stereo rig.
Vargas~\etal \cite{vargas_robust_sensing} proposed robust visual SLAM underwater, leveraging  acoustic, inertial, and altimeter/depth sensors in addition to cameras. Tightly coupled fusion of visual, inertial, and pressure sensors using forward and backward IMU preintegration is discussed in \cite{hu_pressure_fusion}. We use the approach of Rahman~\etal \cite{RahmanIJRR2022} to obtain robust camera pose estimates by fusing visual and inertial
information in real time.
Joshi \etal \cite{JoshiICRA2022} augmented a visual SLAM algorithm 
so that, after loop closures, the map is deformed to preserve the relative pose between each point and its attached keyframes.

Among the few authors who tackle dense underwater stereo, Queiroz-Neto \etal. \cite{underwater_stereo}  modeled light propagation to overcome poor contrast and illumination. 
Wang \etal. \cite{WangICRA2023} presented a pipeline for dense 3D reconstruction based on SVIn2 for visual odometry, real-time stereo matching across frames from a stereo rig and depth map fusion. The synchronized stereo rig is crucial for obtaining consistent and metric depth estimates for each frame.

Recently, 3D modeling, especially for view synthesis, has been addressed by radiance fields, in implicit (Neural Radiance Field - NeRF) \cite{mildenhall2020nerf,luo2024review} or explicit (Gaussian Splatting - GS)  \cite{kerbl20233d,fei2024survey} form.
Their overall success has led to adoption of these techniques in underwater environments. It should be noted that camera poses are externally provided, in general.
Mechanisms to model the water as the medium through which light travels in volumetric rendering have been integrated in NeRF \cite{levy2023seathru,sethuraman2023waternerf,zhang2023beyond,tang2024neural} or GS \cite{liu2024aquatic,yang2024seasplat,zhang2025recgs}.
For instance, SeaThru-NeRF \cite{levy2023seathru} develops a rendering model for scattering media that is able to learn the parameters of the medium along with those of the radiance. 
SeaSplat \cite{yang2024seasplat} extends Gaussian Splatting with the ability to render through a medium, and learns the parameters of the medium, backscatter, and attenuation. %
These methods have achieved impressive novel view synthesis, but geometrically their results are less accurate than conventional Multiple-View Stereo (MVS). To the best of our knowledge, all mentioned underwater radiance fields have been applied to small datasets of dozens of images.

Several complete systems addressing both sparse and dense 3D reconstruction have been deployed.
Kalacska \etal. \cite{kalacska2018freshwater} applied SfM and MVS on aerial and underwater imagery to study freshwater ecosystems. %
Efforts on coral reef monitoring have employed photogrammetric methods underwater \cite{guo2016accuracy,lange2020quick,zhong2023fine}. In a recent publication, Zhong \etal. \cite{zhong2025cuttingedge} reviewed and compared state-of-the-art components for all stages of the photogrammetric pipeline
on a few datasets, each containing a few hundred high-resolution images of coral reefs. 
In a previous research effort that shares many of our objectives, Mahon \etal. \cite{mahon2011reconstructing} presented the design and deployment of a vision-based underwater mapping system to conduct an archaeological survey of the submerged ancient town of Pavlopetri. SLAM followed by dense multi-view reconstruction were used to generate photorealistic 3D models of a large site.

Given the difficulty of mapping an underwater area more than once, it is not surprising that prior underwater multi-session SLAM approaches required strong assumptions. Williams \etal.~\cite{williams2016return} presented work on combining two multibeam surveys of a shipwreck. For both surveys, GPS and water depth measurements were available. Therefore their multi-session SLAM tool focused on fine-tuning the registration between the two dives with the help of SIFT feature matching, for global loop closure. Burguera and Bonin-Font~\cite{burguera2019trajectory} proposed a multi-session mapping framework with visual odometry, also based on SIFT feature matching. However, they initiated each dive at the same spot (at one calibration board) and stayed at a constant altitude. There are also underwater multi-session SLAM approaches for different sensor modalities~\cite{jang2021multi} or for exploiting the geometrical characteristics of the target (e.g., ship hull)~\cite{ozog2016long}.

\section{Proposed Framework}
\subsection{Experimental setup}
Most underwater mapping operations are extremely complex and incur a high time and financial cost. In this work, we introduce a simplified approach that utilizes items commonly carried by scuba divers, such as action cameras and dive computers. The proposed framework provides an easy-to-use, inexpensive approach to 3D reconstruction, comprised of only off-the-shelf action cameras (GoPro\textsuperscript{TM} Hero9 black, or later), a dive computer that can upload the depth profile (Shearwater\textsuperscript{TM} Perdix2 AI is used), and, if the natural illumination is inadequate, additional underwater video lights (two Keldan Video 8XR Ambient 18000lm were deployed; see \fig{fig:beauty} for the two lights mounted on the frame with the two GoPro cameras). In the experiments presented in this paper, two GoPro\textsuperscript{TM} Hero9 black action cameras are rigidly mounted on an aluminum frame with handles, so that their fields of view overlap; see \fig{fig:beauty}. 
Some experiments reported in Section~\ref{sec:results} were performed using unsynchronized videos from both cameras, while other experiments were performed using video from one camera traversing the scene twice.
\invis{Two Keldan\textsuperscript{TM} video lights are attached to the handles providing additional illumination, especially for the interior parts of the shipwreck. A single Shearwater\textsuperscript{TM} Perdix2 AI dive computer (often carried by divers) is used to record the water depth where the diver is, thus recording, approximately, the water depth where the cameras are.} 

The videos from the camera are converted into a ROS bagfile utilizing the framework proposed by Joshi \etal~\cite{JoshiICRA2022,goproROS}. This framework encodes the video (at 30 fps) and the IMU (100 Hz) datastream as ROS messages with synchronized timestamps. The resulting bagfile can be played back and used as input to a number of VIO/VI-SLAM packages; in our case SVIn2~\cite{RahmanIJRR2022} is used. 

Datasets corresponding to three different sessions were recorded in Barbados, at the Pamir shipwreck. With an approximate length of 50m, Pamir rests at a depth of 8m at the bow, and 17m at the seafloor. The first session (data collected in 02/2024) involved two GoPro cameras mounted on a sensor rig with different fields of view; we refer to them as \emph{LeftOnRig} and \emph{RightOnRig}. The second and third session (01/2025) were recorded on the same day, but in two different dives. We refer to these sessions as \emph{Pamir1}, and \emph{Pamir2}. More details are provided in the experiments section.
\invis{
\begin{figure}[t]
     \centering
     {\includegraphics[width=0.9\columnwidth]{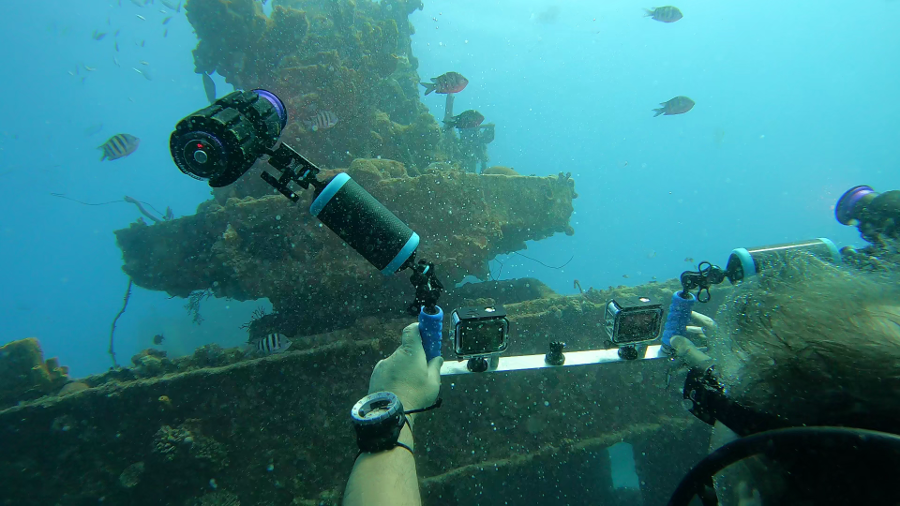}}
     \caption{GoPro setup deployed over the Pamir shipwreck, Barbados during the 2025 deployments.}
     \label{fig:gopro}
 \end{figure}
}

\subsection{Camera Calibration}
While camera calibration above water is a simple matter~\cite{clarke1998development,ZhangIJCV2000,qi2010review}, underwater settings present many challenges as the light passes through more media (water, acrylic enclosure, air, lens, etc.)~\cite{kwon2006effects, roznere2024underwater}. 
We enclose the GoPros in air-tight enclosures with flat-pane windows. It is known that flat-pane windows incur additional light refractions, making cameras incompatible with the traditional perspective projection model. While the Pinax model~\cite{luczynski2017pinax} accurately represents cameras behind flat-pane windows, it requires prior knowledge of the enclosure and water refraction index~\cite{singh2024online}. 
The pinhole camera model with radial-tangential distortion, which is more accessible to implement, can approximate these unique distortions -- as long as the camera is placed as close to the window as possible. This is true in our case, and thus we calibrate the intrinsic parameters based on this model with the help of the calibration boards in \fig{fig:targets}.

\subsection{VI-SLAM: SVIn2}
To produce the keyframes and corresponding trajectory with correct scale, we utilize SVIn2~\cite{RahmanIJRR2022}, a tightly-coupled keyframe-based SLAM system with loop closure, which is able to fuse data from IMU and single or multiple cameras and has been shown to perform well underwater~\cite{JoshiIROS2019}. 

Here we highlight the elements regarding keyframes used in our proposed approach. 
SVIn2 has a frontend that processes each incoming frame at the camera frame rate, performs local optimization for visual-inertial odometry, maintains a bounded window of \emph{keyframes}, and marginalizes states and features which are never used again once they are out of the window, to limit the computation required by the optimization. Keyframes are selected from current frames and old keyframes in the window when the ratio of matched and new keypoints is small. The small ratio corresponds to a significant change in the current scene with some overlap with past scenes. 
The SVIn2 backend performs loop closure and generates a globally optimized trajectory that will be used in our approach: the selected keyframes are added in a pose graph, where each pose gets corrected based on the geometric constraints introduced by the matched descriptors among keyframes. 

\subsection{Absolute Water Depth Correction}
\begin{figure}[h]
    \centering
    \begin{tabular}{lc}
    \begin{subfigure}{0.22\textwidth}
         {\includegraphics[height=0.125\textheight, trim={0.2in, 0in, 0in, 0.3in},clip]{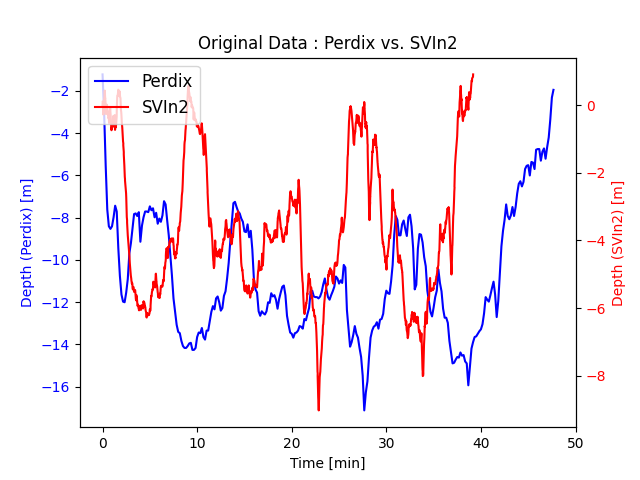}}
         \caption{} \label{depthProcessA}
   \end{subfigure}&
   \begin{subfigure}{0.22\textwidth}
         {\includegraphics[height=0.125\textheight, trim={0.2in, 0in, 0in, 0.3in},clip]{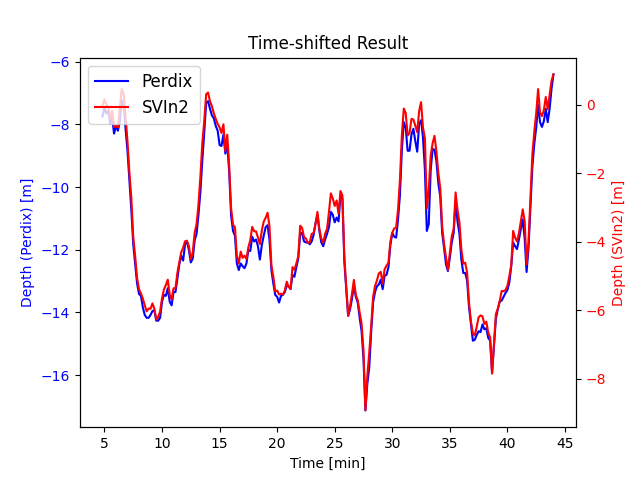}}
         \caption{} \label{depthProcessB}
   \end{subfigure}\\
    \begin{subfigure}{0.22\textwidth}
         {\includegraphics[height=0.125\textheight, trim={0.2in, 0in, 0.5in, 0.3in},clip]{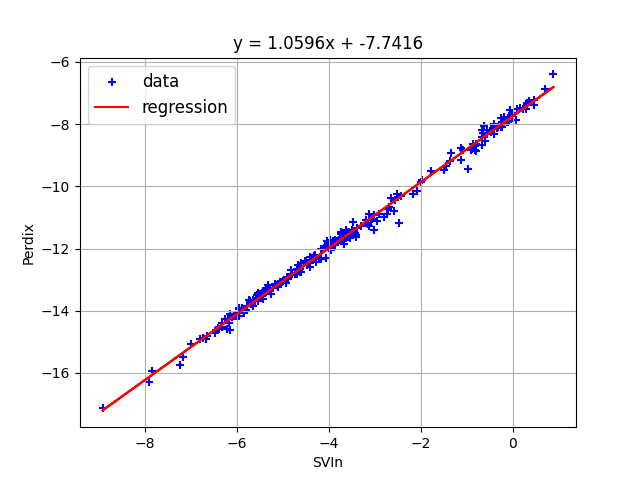}}
         \caption{} \label{depthProcessC}
   \end{subfigure}&
   \begin{subfigure}{0.22\textwidth}
         {\includegraphics[height=0.125\textheight, trim={0.2in, 0in, 0.5in, 0.3in},clip]{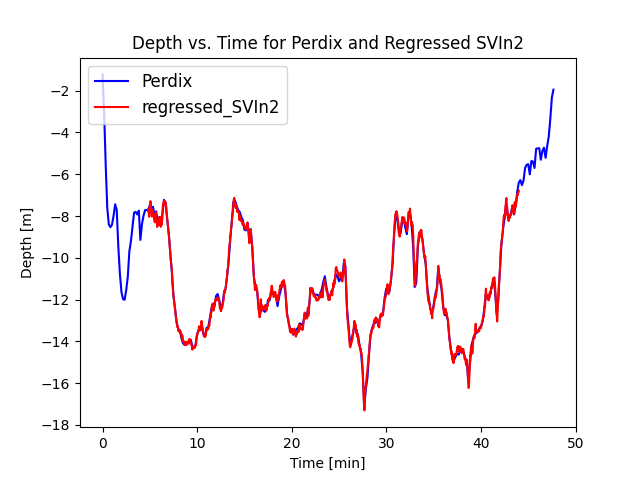}}
         \caption{} \label{depthProcessD}
    \end{subfigure}
    \end{tabular}
    \caption{(a) The SVIn2 z-coordinates (red) and the dive computer depth (blue) before synchronization for the 2024 dataset (LeftOnRig). (b) The two time series shifted to a common time but at different scales. (c) Linear regression between the SVIn2 and the Perdix data. (d) The depth estimates time-shifted and with adjusted depth; in red, the trajectory produced by SVIn2; in blue, the data from the dive computer, for the 2024 LeftOnRig dataset.}
    \label{fig:depthProcess}
\end{figure}

The resulting trajectory from SVIn2 starts at an arbitrary location $[0,0,0]$. As such, while relative motions are correctly recorded, the absolute values are unobservable~\cite{hesch2014camera,yang2019observability}. On the other hand, a dive computer records accurate water depth measurements for the entirety of the dive; see \fig{fig:depthProcess}(a) for the two unsynchronized signals from the 2024 LeftOnRig dataset. Though both signals have approximately the correct time and date, the two clocks often drift, resulting in time differences of several seconds. Furthermore, the dive computer records data at a much slower frequency than SVIn2; in our case Perdix2 AI records every 10 seconds (0.1 Hz) while SVIn2 produces data at different frequencies; 1.57 Hz and 1.89 Hz, for 2024 and  2.3 Hz, and 3.4 Hz for the 2025 datasets, depending on the selected keyframes. 

For each dataset, a common time period is selected, and both time series are interpolated to 100 Hz, such that the two signals have the same frequency. The cross-correlation between the two signals (SVIn2 and Perdix) was used to estimate the time shift; see \fig{fig:depthProcess}(b) where the two signals from the 2024 LeftOnRig dataset are converted to Perdix time, while still in different scales (Perdix blue, and SVIn2 red). The next step is linear regression to calculate the shift in values; see \fig{fig:depthProcess}(c) for the scatter plot of all the data and the results of linear regression. Finally, the SVIn2 time series of the z-coordinate is transformed to the real depth; see \fig{fig:depthProcess}(d) for the two signals converted in the same time and depth for the 2024 LeftOnRig dataset. The 2024 RightOnRig dataset shares the same depth profile and the plots are omitted. The same process is applied to the Pamir1 (see \fig{fig:depthCorr} left) and Pamir2 (see \fig{fig:depthCorr} right) datasets. For the three sessions, the GoPro time series was shifted 289.8 sec, 25.94 sec, and 29.91 sec respectively. The depth offsets were 7.74, 8.11, and 10.95 meters, respectively. These results are consistent with the experiments, as the GoPro started for the two first datasets (2024 Left and Right on Rig and Pamir1) at the bow of the wreck, which is shallower, while the third dataset (Pamir2) was started at the middle of the ship, which rests in deeper water. 

\begin{figure}[h]
    \centering
    \begin{tabular}{lc}
         \hspace{-0.05in}{\includegraphics[height=0.13\textheight, trim={0.2in, 0in, 0.5in, 0.3in},clip]{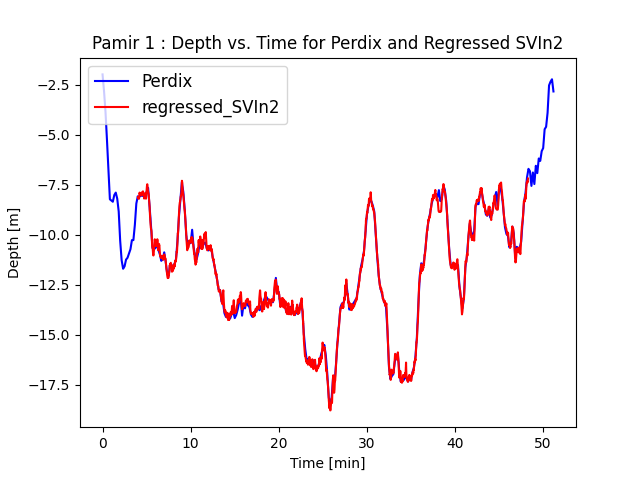}}&
         \hspace{-0.05in}{\includegraphics[height=0.13\textheight, trim={0.2in, 0in, 0.5in, 0.3in},clip]{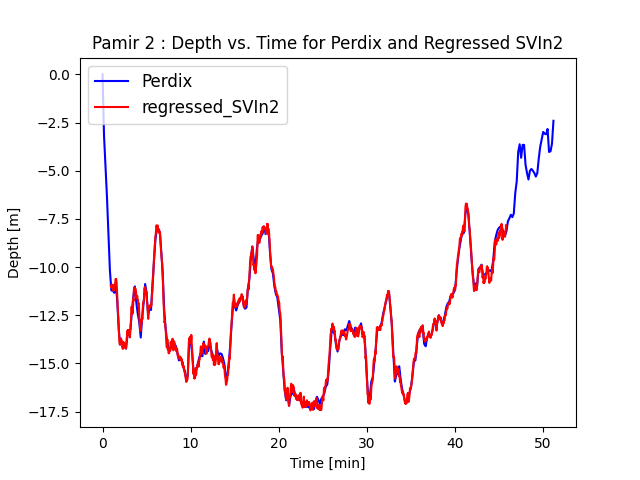}}
    \end{tabular}
    \caption{The depth estimates for the other two datasets (Pamir1 and Pamir2), time-shifted and depth aligned, in red is the trajectory produced by SVIn2, in blue are the data from the dive computer.}
    \label{fig:depthCorr}
\end{figure}

\invis{

\begin{itemize}
    \item Truncate the two time series using only the common time between them.
    \item Resample both to have data at the same frequency.
    \item Calculate the time shift from correlation.
    \item Shift SVIn2 series accordingly.
    \item Linear interpolation between SVIn2 and the complete Perdix time series to Estimate the SVIn2 value at each Perdix Value.
    \item Linear fit between the two data series. 
    \item Estimate the linear transformation between Perdix and SVIn2.
    \item Transform the SVIn2 Trajectory.
\end{itemize}  
\begin{figure}[h]
    \centering
    \begin{tabular}{lcc}
         \hspace{-0.1in}{\includegraphics[height=0.091\textheight, trim={0.2in, 0in, 0.5in, 0.3in},clip]{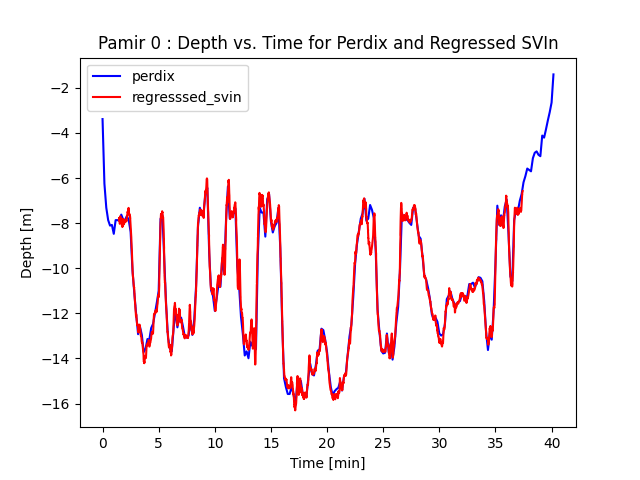}}&
         \hspace{-0.15in}{\includegraphics[height=0.091\textheight, trim={0.0in, 0in, 0.5in, 0.3in},clip]{figures/Pamir1_final.png}}&
         \hspace{-0.14in}{\includegraphics[height=0.091\textheight, trim={0.2in, 0in, 0.5in, 0.3in},clip]{figures/Pamir2_final.png}}
    \end{tabular}
    \caption{The depth estimates time-shifted for the three deployment sessions. In red the SVIn2 produced trajectory and in blue the data from the dive computer.}
    \label{fig:depthCorr}
\end{figure}
}

\subsection{Trajectory Correlation}\label{sec:traj_corr}

\begin{figure}[h]
    \centering
    \leavevmode
    \begin{tabular}{lc}
         {\includegraphics[height=0.09\textheight]{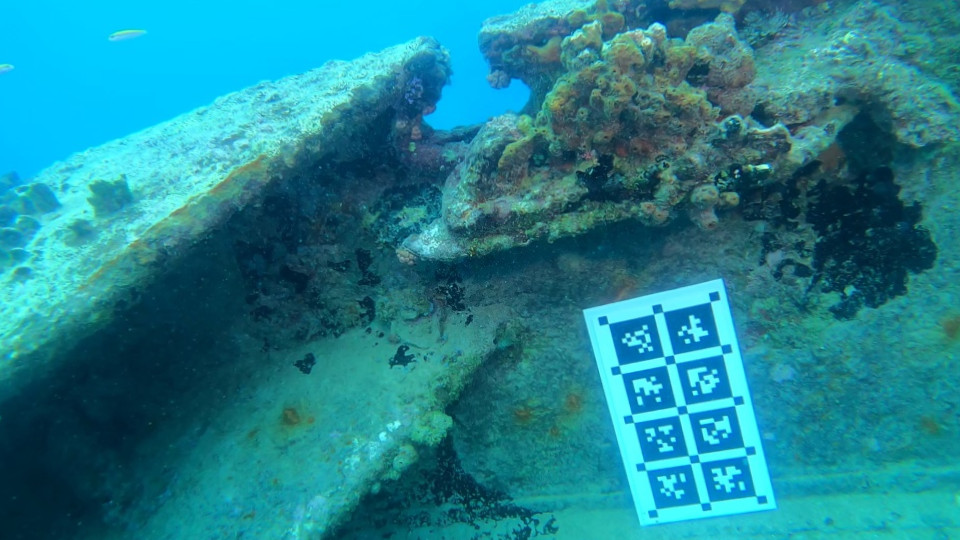}}&
         {\includegraphics[height=0.09\textheight]{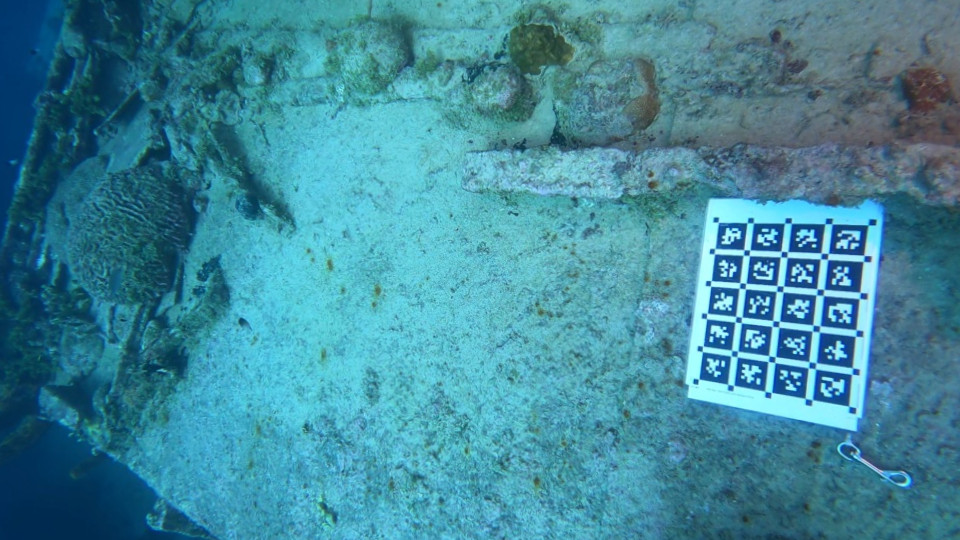}}
         \end{tabular}
    \caption{The 4-by-2 and the 5-by-4 targets at Pamir wreck, which stayed at a fixed location for the Pamir1 and Pamir2 sessions. }
    \label{fig:targets}
\end{figure}

To transform the two trajectories into a unified coordinate frame, we utilize a calibration target placed at the bow, where deployment starts.
\fig{fig:targets} shows the two targets on the Pamir wreck. 
The \textit{aprilgrid\footnote{\href{https://github.com/powei-lin/aprilgrid}{https://github.com/powei-lin/aprilgrid}}} package is employed to detect the calibration targets.
For each observation, we compute the pose of the target in the global (world) frame. Let $T_{C_i}^{W}$ be the pose of camera $i$ in the world frame $W$, and $T_{m}^{C_i}$ be the relative pose of the marker $m$ with respect to camera $i$. The pose of the marker in the world frame, $T_{m,i}^{W}$, is derived as:
\begin{equation}
T_{m,i}^{W} = T_{C_i}^{W} T_{m}^{C_i}    
\end{equation}

First, we estimate the average pose of the calibration target. Because the calibration target consists of multiple individual tags, we perform pose estimation for each constituent tag and average their translations and orientations. Let $t_k \in \mathbb{R}^3$ be the translation of the $k$-th tag, and $\Theta_k = [\phi_k, \theta_k, \psi_k]^T$ be its orientation expressed as roll, pitch, and yaw angles. The average position $\bar{t}$ and orientation $\bar{\Theta}$ over $N$ tags are calculated as:
\begin{equation}
\bar{t} = \frac{1}{N} \sum_{k=1}^{N} t_k, \quad \bar{\Theta} = \frac{1}{N} \sum_{k=1}^{N} \Theta_k
\end{equation}

\noindent Despite the theoretical risks of gimbal lock associated with Euler angle averaging, the close proximity of our candidate poses ensured a stable global transformation. In cases of higher rotational uncertainty, mapping the rotations to a quaternion space would provide a more generalized solution. 

Next, we filter outlier observations for each target across all camera frames. Let $t_{m,i}$ be the translation of target $m$ estimated from camera $i$, and $\bar{t}_m$ be the mean translation of that target across all observations. The Euclidean distance $D_i$ for each observation is calculated as:
\begin{equation}
D_i = \| t_{m,i} - \bar{t}_m \|_2
\end{equation}
We compute the mean $\mu_D$ and standard deviation $\sigma_D$ of these distances. Any target estimate with a distance $D_i$ falling outside one standard deviation from the mean is considered an outlier and is removed.
Subsequently, we filter the remaining inliers based on their orientation estimates, which are typically more susceptible to noise. From the rotation matrix of each pose, we extract the principal axes angles $[\phi_i, \theta_i, \psi_i]^T$. We compute the mean and standard deviation for each angle component. An estimate is removed if any of its roll, pitch, or yaw angles fall outside one standard deviation from their corresponding mean. An updated, refined average pose $\bar{T}_{m}^{W}$ is then recalculated using only this strict inlier set.

To align two distinct trajectories (e.g., Trajectory A and Trajectory B), we leverage the refined poses of the same physical target observed in both coordinate systems. Let $T_{m}^{A}$ and $T_{m}^{B}$ be the estimated poses of the target in World Frame A and World Frame B, respectively. The transformation $T_{B}^{A}$ that maps coordinates from Frame B into Frame A is computed as:
\begin{equation}
T_{B}^{A} = T_{m}^{A} (T_{m}^{B})^{-1}
\end{equation}
This calculation is performed independently for each of the two calibration targets, yielding a set of candidate transformations $\{T_{B, j}^{A}\}_{j=1}^{M}$ between the two trajectories. Finally, to establish a single global transformation to apply to the entirety of Trajectory B, we average these candidate transformations. We decouple the translation and rotation components of each candidate transformation $T_{B, j}^{A}$. The rotation matrices are converted into roll, pitch, and yaw angles, denoted as $\Theta_j = [\phi_j, \theta_j, \psi_j]^T$, and the translation components are extracted as $t_j$. The final global translation $\bar{t}_{\text{global}}$ and global orientation $\bar{\Theta}_{\text{global}}$ are then calculated as the arithmetic means of these components:
\begin{equation}
\bar{t}_{\text{global}} = \frac{1}{M} \sum_{j=1}^{M} t_j, \quad \bar{\Theta}_{\text{global}} = \frac{1}{M} \sum_{j=1}^{M} \Theta_j
\end{equation}
These averaged translation and orientation values are recombined into a single global transformation matrix, which is applied to move the entirety of Trajectory B into the coordinate frame of Trajectory A.

\begin{figure*}[h]
    \centering
    \leavevmode
    \begin{tabular}{lcc}
         {\includegraphics[height=0.2\textheight, trim={7in, 1in, 5.7in, 0.5in},clip]{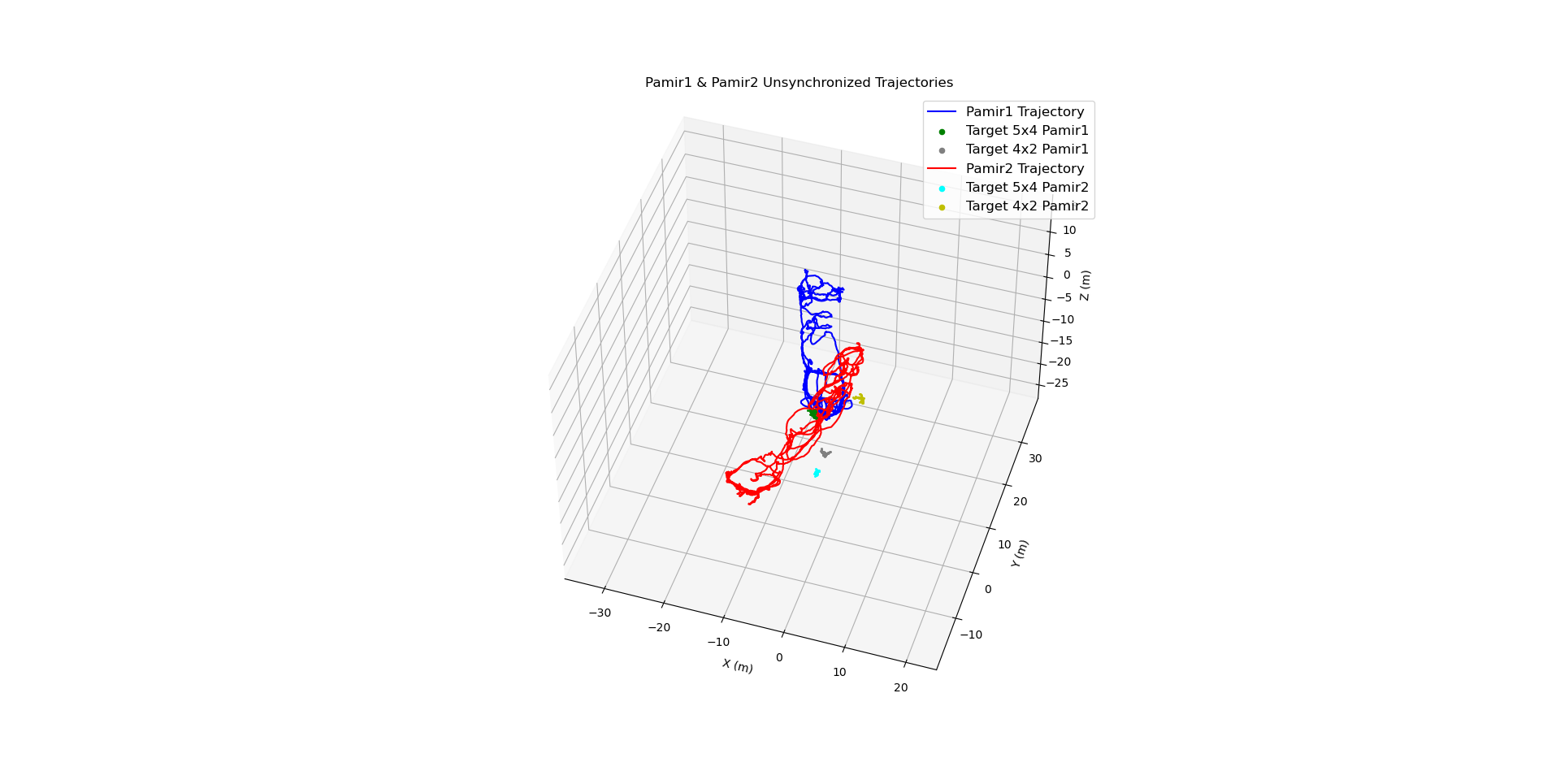}\label{fig:trajSyncA}}&
         {\includegraphics[height=0.2\textheight, trim={6.7in, 2.2in, 5.7in, 0.5in},clip]{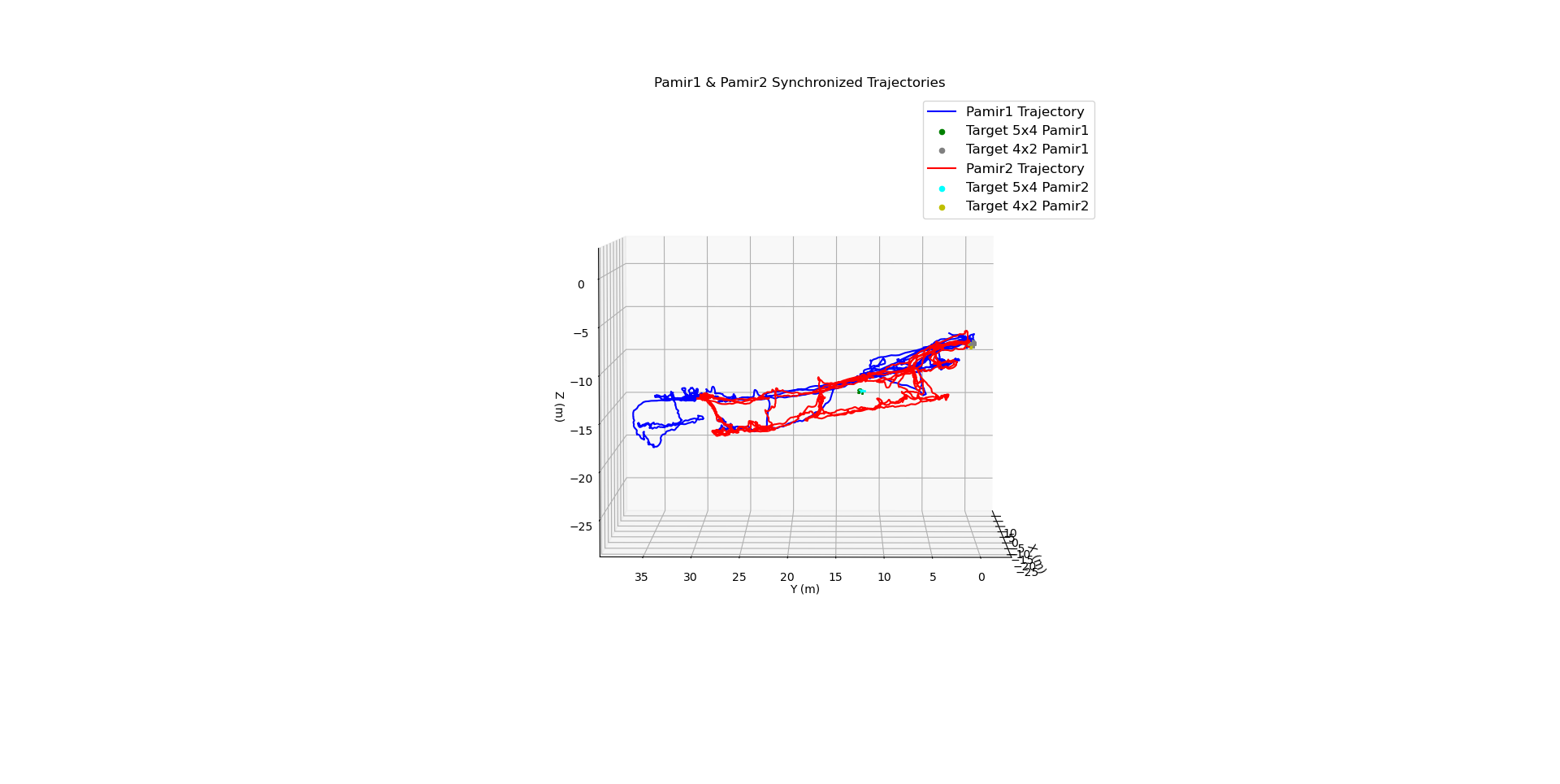}\label{fig:trajSyncB}}&
         {\includegraphics[height=0.2\textheight, trim={6.7in, 1.8in, 5.7in, 0.5in},clip]{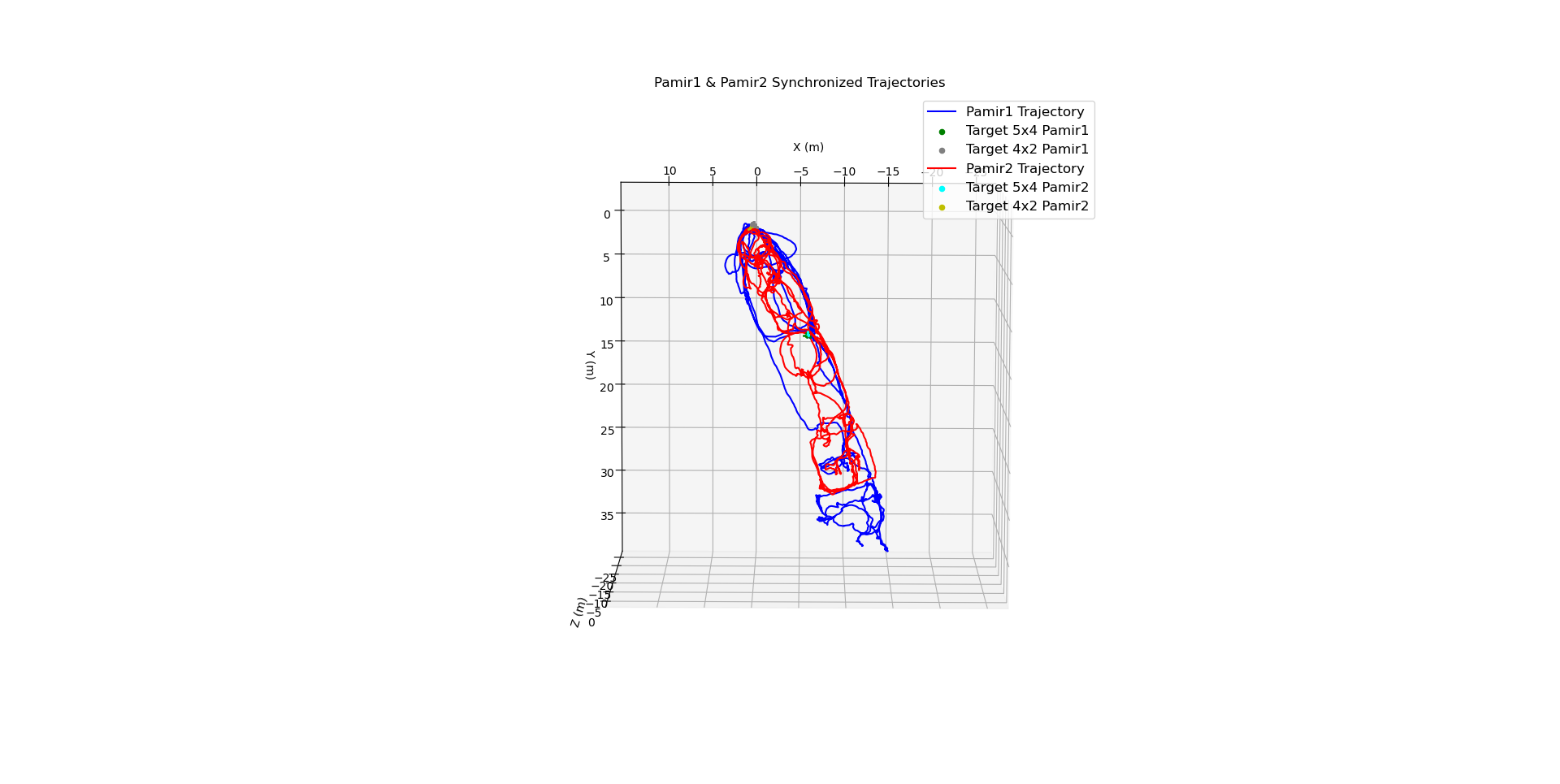}\label{fig:trajSyncC}}
    \end{tabular}
    \caption{The left figure presents the second and third trajectory as they were produced by SVIn2; please note that each trajectory starts at [0,0,0]. Both targets detected are also plotted for each session relative to their corresponding trajectories. The middle and right figures present the two trajectories brought to the same reference frame and the depth (z-axis) is adjusted based on the dive computer data. Both targets appear in the same place and the outline of the wreck is clearly visible.}
    \label{fig:trajSync}\vspace{-1em}
\end{figure*}

\subsection{Global Bundle Adjustment}

To improve the geometric consistency of the camera poses and to prepare the data for dense reconstruction, we use the sparse COLMAP pipeline \cite{colmap_sfm} on the \textit{keyframes} of SVIn2. 

Unlike SVIn2, COLMAP is a Structure-from-Motion system designed for collections of images, which have not necessarily been acquired sequentially or even by the same camera. This makes it possible to link images across trajectories by establishing feature correspondences between them, even if these trajectories do not contain common fiducial markers such as those in Section~\ref{sec:traj_corr} above. Natural features observed in the images are sufficient.

COLMAP uses SIFT features \cite{lowe2004distinctive} to detect pairwise pose constraints among the images, consolidates these constraints, and extracts the final camera poses by refining the poses generated by SVIn2. Specifically, we use the sparse modules of COLMAP as follows:
\begin{itemize}
    \item Extract SIFT features from the images of two or more trajectories;
    \item Detect corresponding features across all image pairs;
    \item Initialize the poses of the cameras using poses from the aligned SVIn2 trajectories;
    \item Iteratively reduce reprojection error via bundle adjustment.
\end{itemize}

This approach benefits from the strengths of both the VIO and the SfM systems: it operates in metric space with the scale provided by visual-inertial odometry and the depth sensor; it avoids processing all frames with COLMAP, which would be prohibitively costly in compute and memory usage. Moreover, COLMAP receives frames that were selected as keyframes instead of a decimated sequence; it improves the overall precision of camera poses and reconstructed features via bundle adjustment. Empirically, we have observed that running COLMAP on the keyframes yielded breaks in the trajectory due to tracking difficulties. On the other hand, using the poses from SVIn2 is more robust due to the availability of the IMU signals.
(Note that we are not able to use GLOMAP \cite{pan2024glomap}, a recent extension of COLMAP, because it does not support refinement of externally provided initial poses.)

\subsection{Dense Reconstruction}
COLMAP's dense reconstruction pipeline \cite{schonberger16_dense} consists of three steps: depth map estimation using patch-match stereo with per-pixel view selection, depth map fusion, and mesh generation. 
The first step estimates a depth per pixel using each input image with a camera pose as reference, and selecting target images for each pixel according to geometric criteria. The output is a set of depth maps, which 
are further improved by the subsequent fusion step.
Finally, a triangular mesh can be obtained either by solving a Poisson problem \cite{kazhdan2013screened}, which computes an indicator function signifying which parts of space are inside or outside the surface.

\begin{figure}[t]
    \centering
    \begin{tabular}{c}
         \fbox{\includegraphics[width=0.45\textwidth, trim={0.0in, 0in, 0in, 0in},clip]{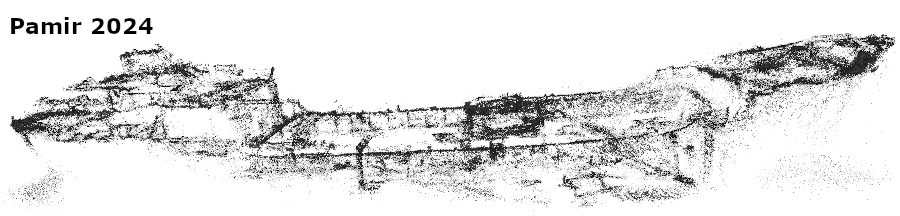}}\\
         \fbox{\includegraphics[width=0.45\textwidth, trim={0.4in, 2.5in, 0.5in, 2.6in},clip]{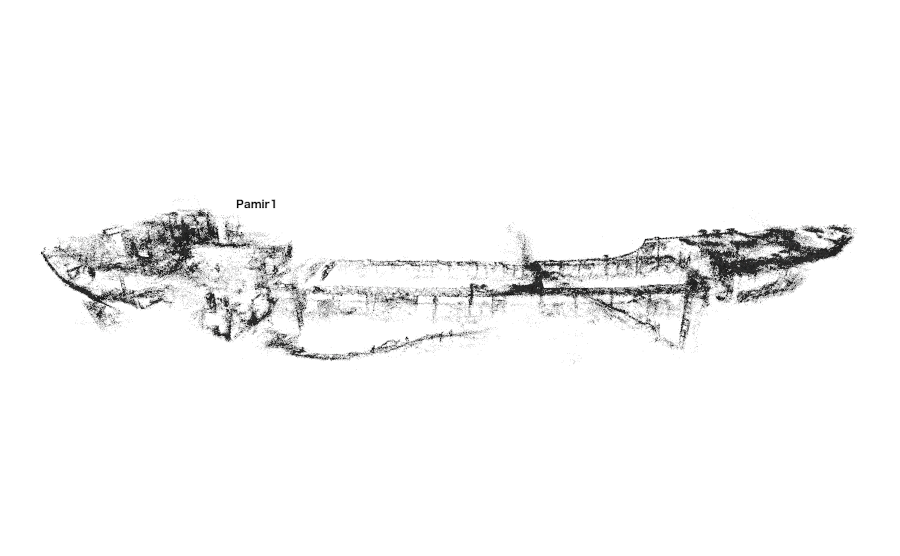}}\\
         \fbox{\includegraphics[width=0.45\textwidth, trim={0.4in, 2.8in, 0.5in, 2.2in},clip]{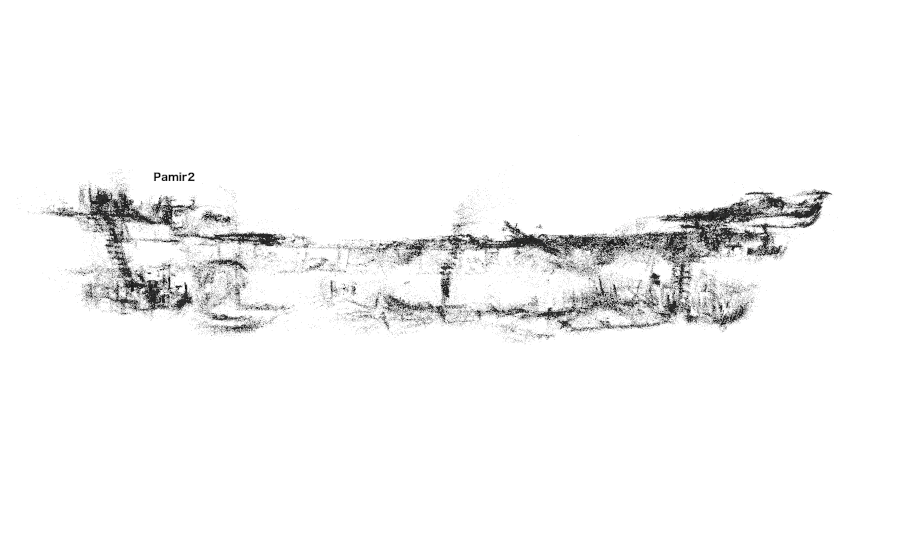}}
    \end{tabular}
    \caption{The sparse reconstruction from SVIn2 for the three sessions. Different parts of the wreck were mapped, including the interior sections in the bow and the stern of the wreck.}
    \label{fig:SvinReconstruction}
    \vspace{-2em}
\end{figure}

\section{Experimental Results}\label{sec:results}
\subsection{Datasets}
\begin{figure}[h]
     \centering
     {\includegraphics[width=0.85\columnwidth]{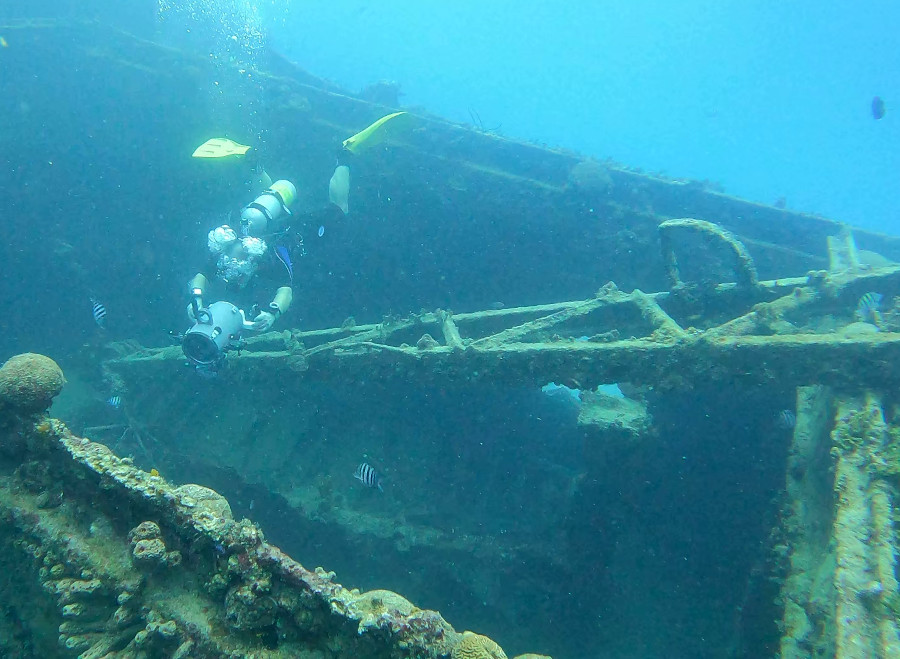}}
     \caption{2024 deployment at the Pamir shipwreck, Barbados. The sensor rig with the two GoPros mounted to the left and right.}
     \label{fig:Pamir2024}
 \end{figure}

Three dives were performed at the Pamir shipwreck with durations of 55 (2024), 51 and 51 (2025) minutes respectively, while data were recorded for 39, 44, and 45 minutes respectively. In the first dive, the cameras were attached to a sensor rig; see \fig{fig:Pamir2024} for a photo of the setup. For the duration of the second and third dives, two calibration targets (4-by-2 and 5-by-4) were fixed in separate locations. The application of SVIn2 to the three sessions resulted in \num{3699} and \num{4,464} keyframes for the left and right cameras of the 2024 session and \num{5402}, and \num{8285} keyframes for the two 2025 sessions; a selection of \num{21850} keyframes out of a total of approximately \num{300600} frames. The datasets are available online\footnote{\url{https://github.com/AutonomousFieldRoboticsLab/Pamir_Visual_Inertial_Dataset}}.

\subsection{Sparse reconstruction and Camera Trajectory Generation}
SVIn2 produces a consistent sparse representation for each of the three sessions. As can be seen in \fig{fig:SvinReconstruction}, the wreck is reconstructed where the camera passed. However, coverage remains incomplete. Mapping a large, structure such as a wreck, requires several dive sessions.

\paragraph{Single Session (Pamir1)} The camera positions from SVIn2 were used  as ``GPS'' initialization points for COLMAP. As can be seen in \fig{fig:Pamir1} the camera positions from SVIn2 (in red) have been refined to the COLMAP final camera positions (in yellow) to satisfy the global optimization constraints. The resulting reconstruction is also presented in the same image. 
\invis{
\begin{figure}[h]
    \centering
    \leavevmode
    \begin{tabular}{lc}
         \hspace{-0.1in}{\includegraphics[width=0.24\textwidth]{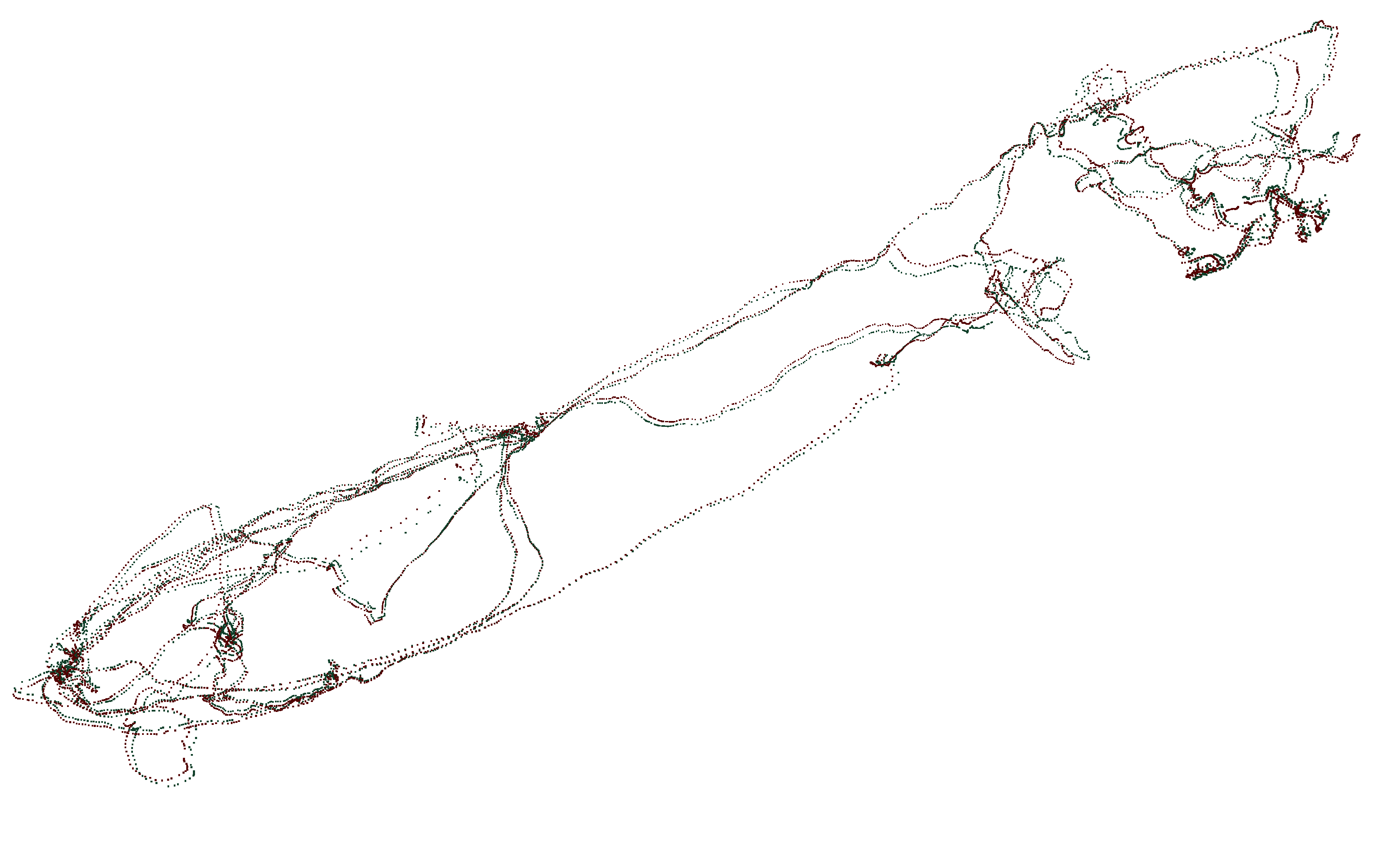}}&
        \hspace{-0.19in}{\includegraphics[width=0.24\textwidth, trim={0in, 0in, 0in, 0in},clip]{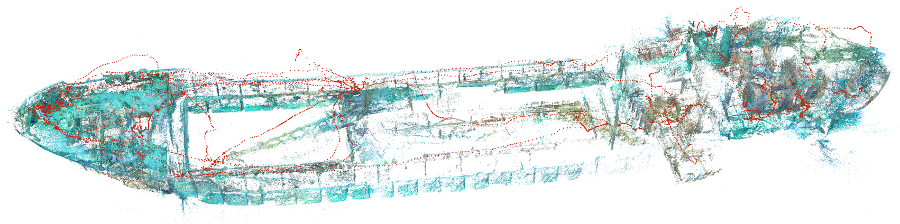}}
         \end{tabular}
    \caption{The trajectory from SVIn2 (red) and the refined trajectory from COLMAP (green) are shown on the left. The sparse reconstruction (with the camera poses in red) is shown on the right.}
    \label{fig:Pamir1}
\end{figure}
}
\begin{figure}[h]
     \centering
     {\includegraphics[width=0.8\columnwidth, trim={1.0in, 1.0in, 1.0in, 1.0in},clip]{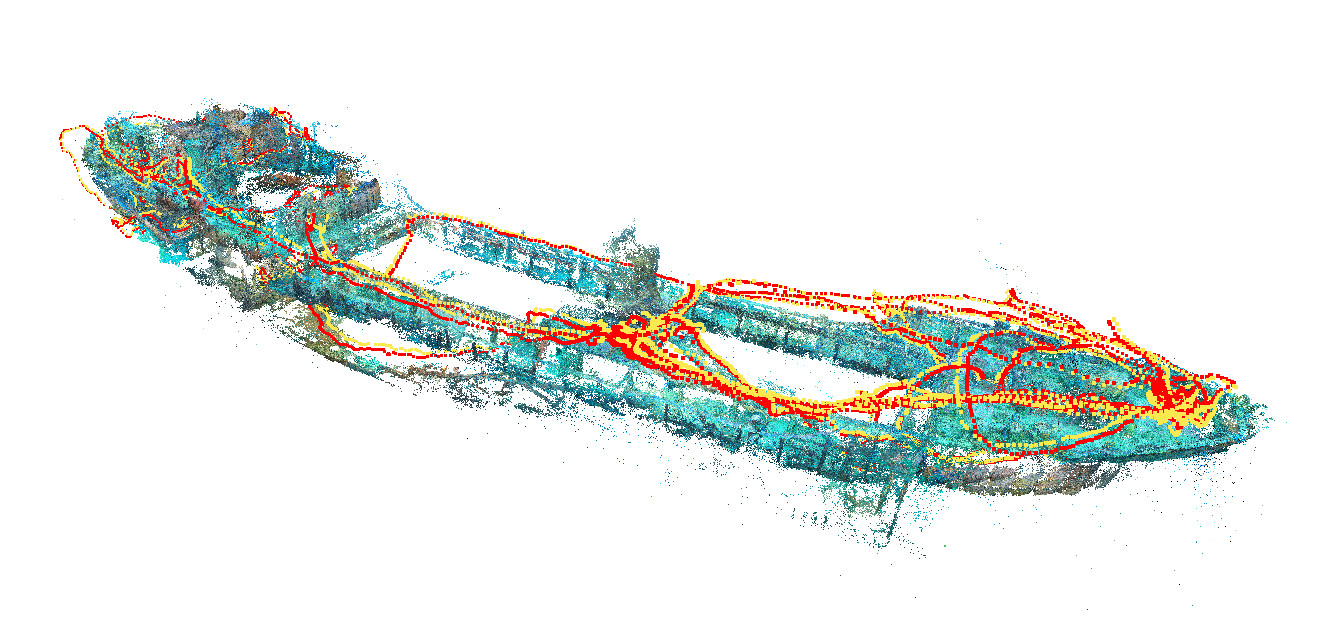}}
     \caption{The trajectory from SVIn2 (red) and the refined trajectory from COLMAP (yellow) are shown over the  sparse reconstruction of the Pamir shipwreck, Barbados.}
     \label{fig:Pamir1}
     \vspace{-1em}
 \end{figure}

\begin{figure*}[h]
    \centering
    \begin{tabular}{ll}
\invis{    \hspace{-0.05in}\begin{subfigure}{0.45\textwidth}
         \fbox{\includegraphics[width=0.95\textwidth, trim={0in, 4in, 0in, 0in},clip]{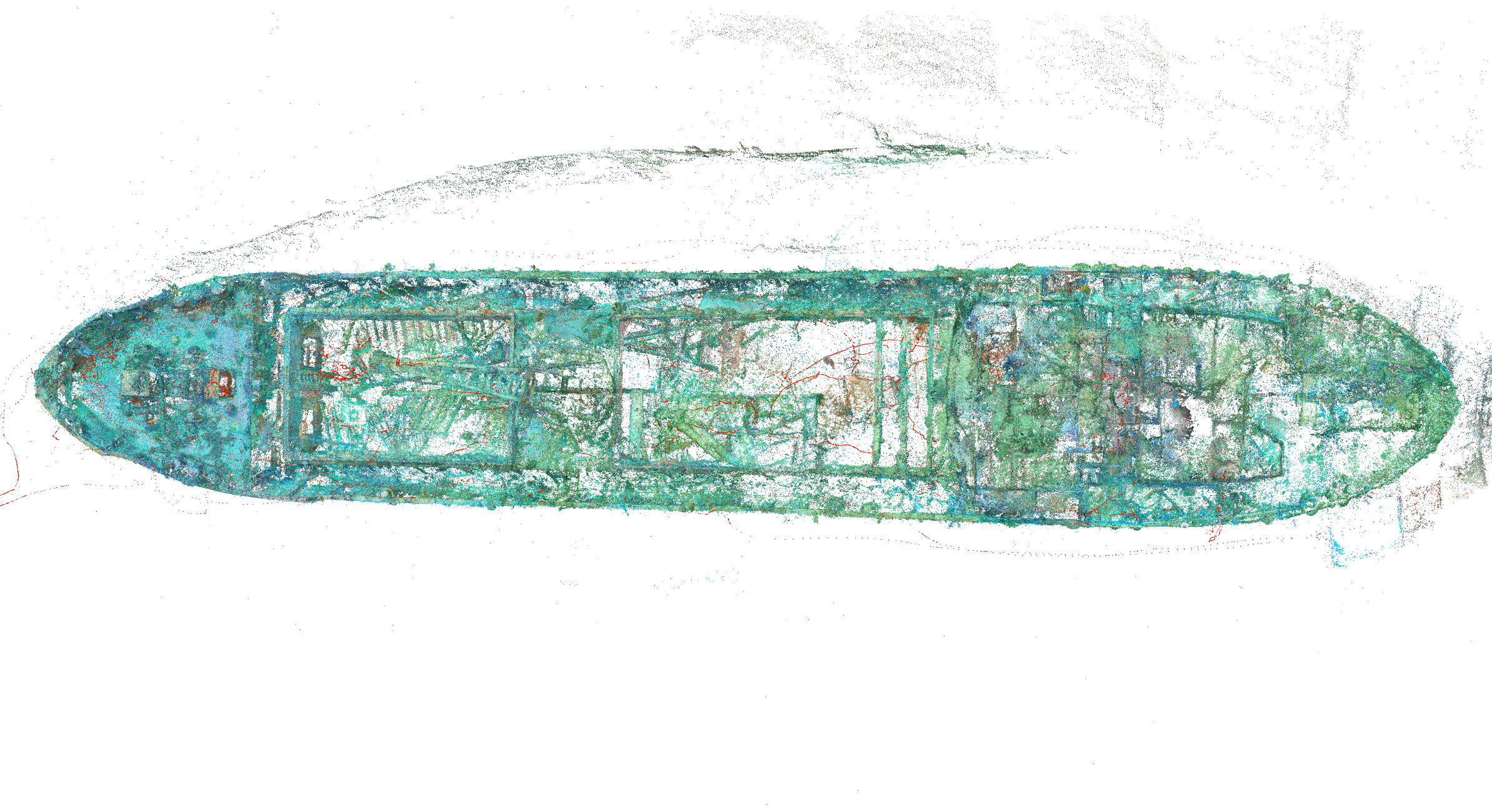}}
         \caption{} \label{SparsePartsA}
   \end{subfigure}&
   \hspace{-0.1in}\begin{subfigure}{0.45\textwidth}
          \fbox{\includegraphics[width=0.95\textwidth, trim={0in, 4in, 0in, 0in},clip]{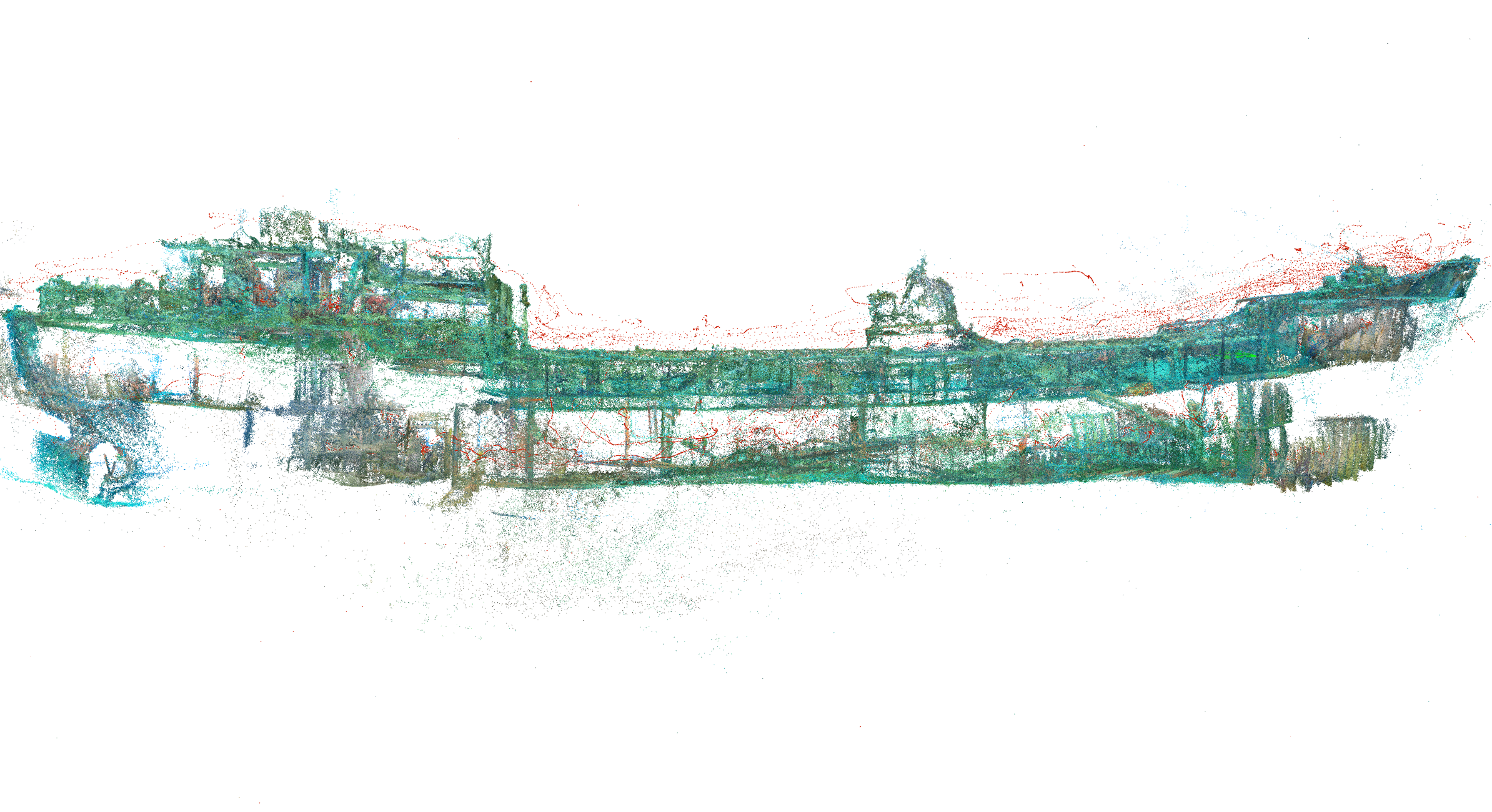}}
         \caption{} \label{SparsePartsB}
   \end{subfigure}\\}
   \hspace{-0.05in}\begin{subfigure}{0.45\textwidth}
         \fbox{\includegraphics[width=0.95\textwidth, trim={0in, 0in, 0in, 0in},clip]{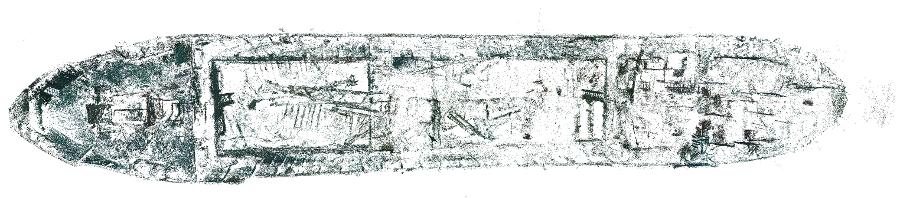}}
         \caption{} \label{SparseAllA}
   \end{subfigure}&
   \hspace{-0.1in}\begin{subfigure}{0.45\textwidth}
          \fbox{\includegraphics[width=0.95\textwidth, trim={0in, 0in, 0in, 0in},clip]{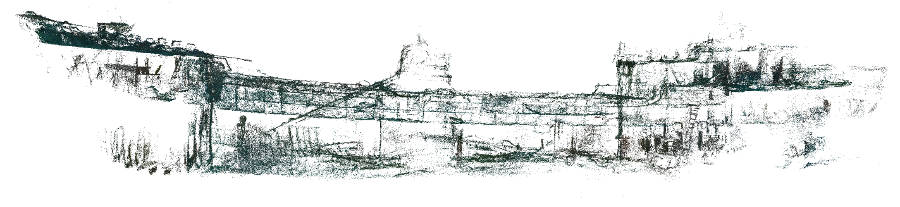}}
         \caption{} \label{SparseAllB}
   \end{subfigure}
    \end{tabular}
    \vspace{-0.1in}\caption{Full sparse reconstruction of the Pamir shipwreck from the two 2025 sessions. (a) Top view. (b) Side view of the wreck.}
    \label{fig:SparseAll}
\end{figure*}

\begin{figure*}[h]
    \centering
    \begin{tabular}{lc}
    \hspace{-0.05in}\begin{subfigure}{0.25\textwidth}
         \fbox{\includegraphics[width=0.95\textwidth, trim={0in, 0in, 0in, 0in},clip]{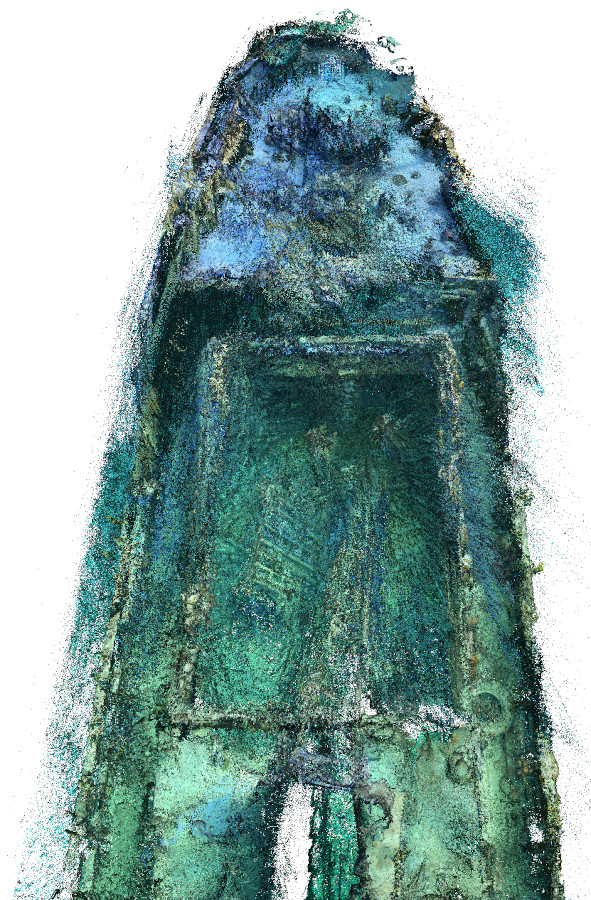}}
         \caption{} \label{SparsePartsA}
   \end{subfigure}&
   \hspace{-0.1in}\begin{subfigure}{0.62\textwidth}
          \fbox{\includegraphics[width=0.95\textwidth]{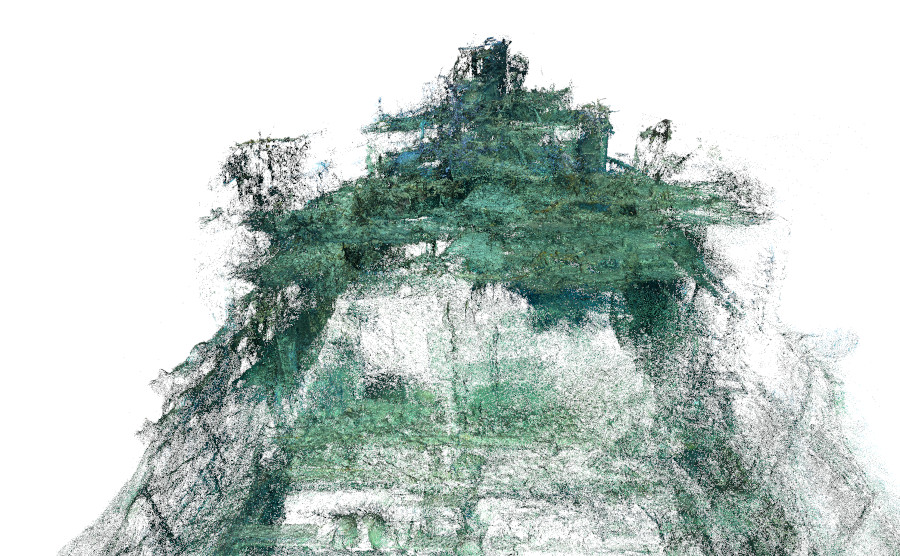}}
         \caption{} \label{SparsePartsB}
   \end{subfigure}\\
    \end{tabular}
    \begin{tabular}{lcc}
   \hspace{-0.05in}\begin{subfigure}{0.32\textwidth}
         \fbox{\includegraphics[width=0.95\textwidth]{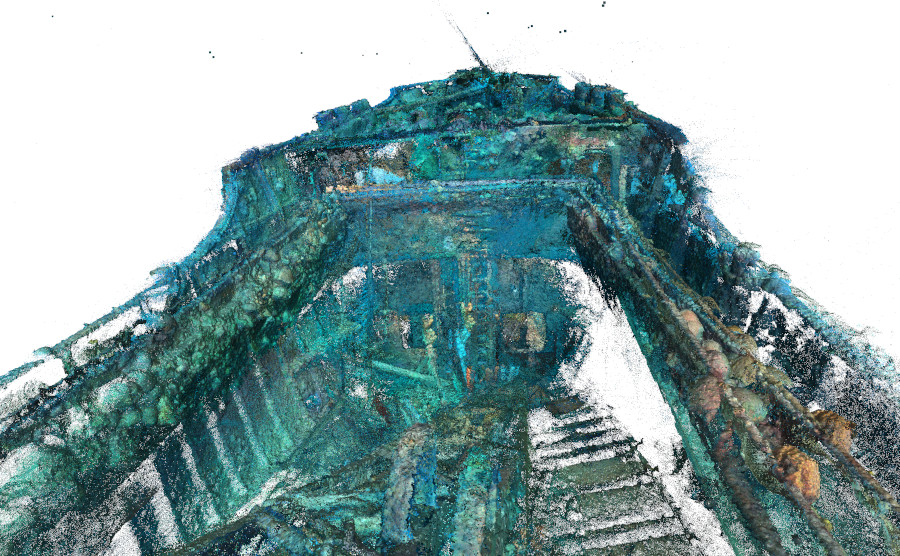}}
         \caption{} \label{SparsePartsD}
   \end{subfigure}&
   \hspace{-0.1in}\begin{subfigure}{0.32\textwidth}
          \fbox{\includegraphics[width=0.95\textwidth]{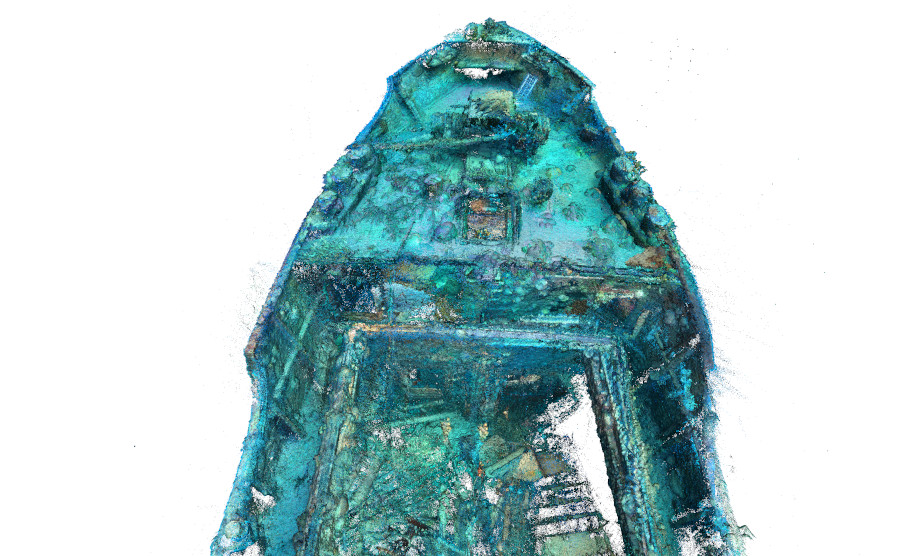}}
         \caption{} \label{SparsePartsE}
   \end{subfigure}&
    \hspace{-0.1in}\begin{subfigure}{0.32\textwidth}
          \fbox{\includegraphics[width=0.95\textwidth]{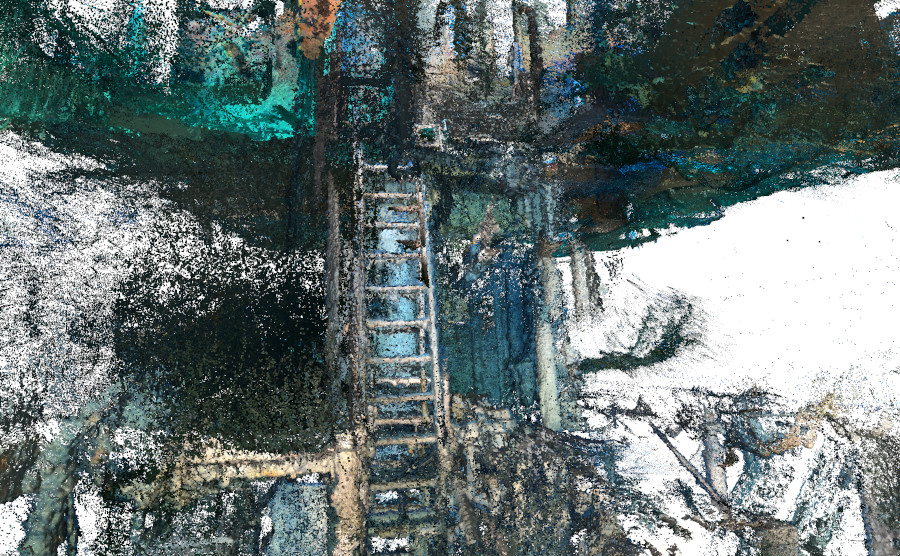}}
         \caption{} \label{SparsePartsF}
   \end{subfigure}
    \end{tabular}
    \vspace{-0.1in}\caption{Segments of the dense reconstruction from COLMAP presenting details of the mapped structure. (a) 2024: Top view showing the cargo hold crane and target (front right). (b) 2024: Stern of the wreck. (c) 2025: Perspective of the bow and cargo hold. (d) 2025: Bow view with visible target. (e) 2025: Interior view of the engine room.
    }
    \label{fig:SparseParts}
    \vspace{-1em}
\end{figure*}

\paragraph{Multi-Session Reconstruction} As can be seen in \fig{fig:SparseAll}, utilizing the two  sessions of 2025 resulted in more area covered, including several parts of the ship's interior.  In \fig{fig:SparseParts} details of the reconstruction can be seen. Figures \ref{fig:SparseParts}(a-b) are from the 2024 deployment and (c-e) from 2025. The bow and stern of the wreck are presented on the top row using the 2024 datasets. The second row presents data from the 2025 deployment, including an interior view of the engine room (Fig. \ref{fig:SparseParts}(e)). In Fig. \ref{fig:SparseParts}(d), the target is visible at the bow of the ship. 

\invis{
present outside views of the wreck, while \fig{fig:SparseParts}(d,e) present interior views. The square opening in the middle of the bow (with several red camera poses inside) leads into an inside room accessible via the ladder visible in \fig{fig:SparseParts}(e). The engine room in \fig{fig:SparseParts}(d) is a cluttered space with a vertical ladder at the back leading to the corridor going across the rear part of the wreck seen in \fig{fig:SparseParts}(b); traversals through that corridor resulted in several red camera poses clustered in the middle of the image. During the mapping of this wreck more time was spend exploring the interior structures and thus the outside part of the wreck is less well mapped. 
}

\subsection{Dense Reconstruction}

\begin{figure}[t]
     \centering
     {\includegraphics[width=0.9\columnwidth]{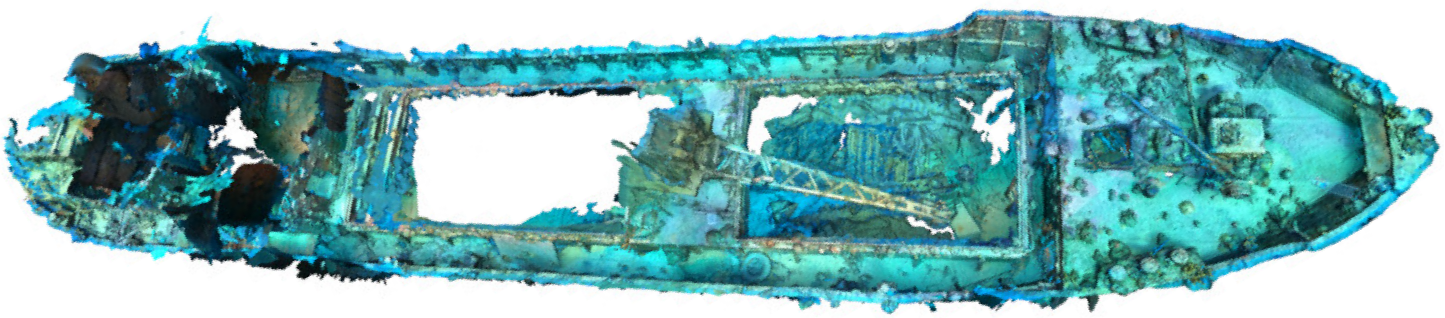}}\\
    \caption{Poisson mesh of selected frames from Pamir1.}
     \label{fig:poisson}
 \end{figure}

\begin{figure*}[h]
    \centering
    \begin{tabular}{lcc} \invis{
    \hspace{-0.05in}\begin{subfigure}{0.3\textwidth}
         \fbox{\includegraphics[width=0.95\textwidth]{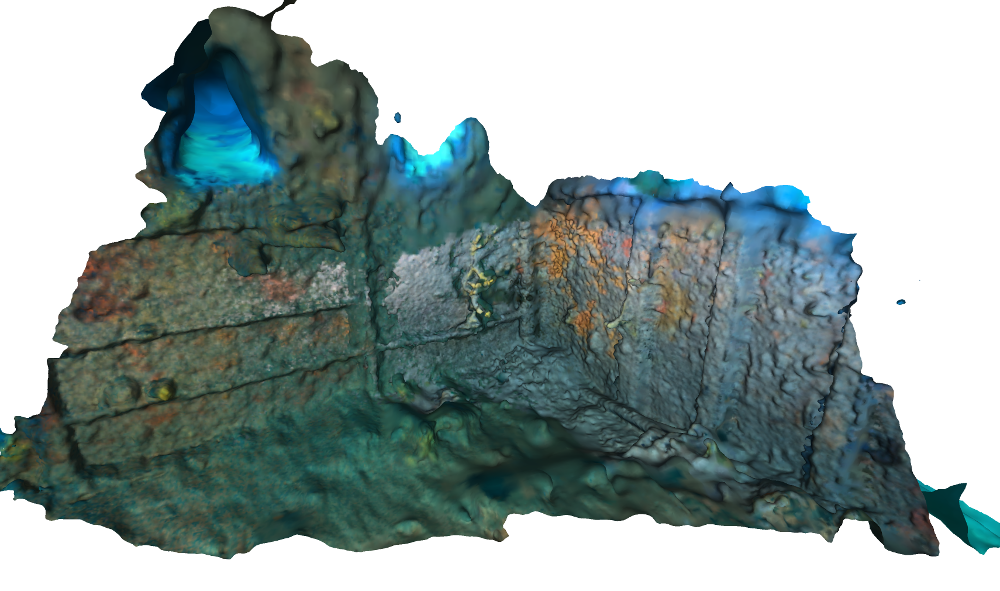}}
         \caption{} \label{DensePartsA}
   \end{subfigure}&
   \hspace{-0.1in}\begin{subfigure}{0.3\textwidth}
          \fbox{\includegraphics[width=0.95\textwidth]{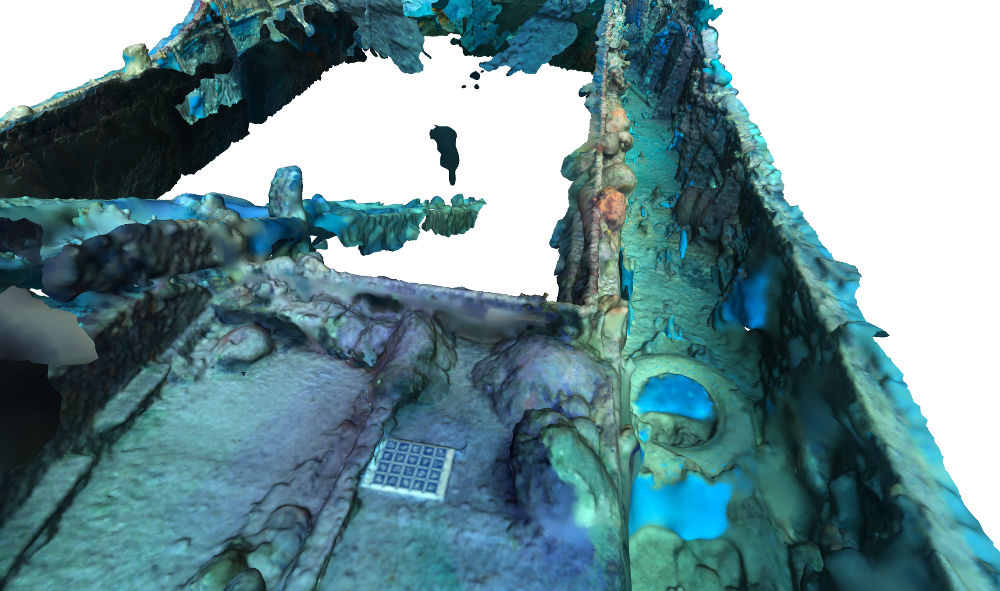}}
         \caption{} \label{DensePartsB}
   \end{subfigure}&
    \hspace{-0.1in}\begin{subfigure}{0.25\textwidth}
          \fbox{\includegraphics[width=0.95\textwidth, trim={0in, 0in, 0in, 0in},clip]{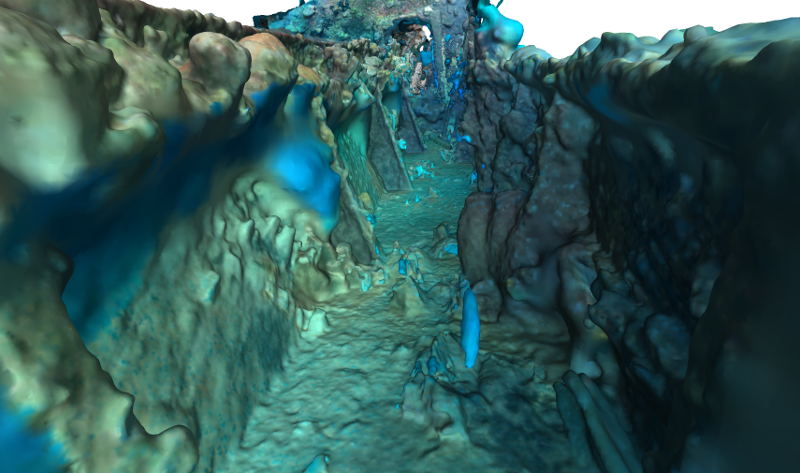}}
         \caption{} \label{DensePartsC}
   \end{subfigure}\\}
    \hspace{-0.4in}\begin{subfigure}{0.2\textwidth}
         \fbox{\includegraphics[height=0.14\textheight, trim={0.0in, 0.0in, 0.0in, 0.0in},clip]{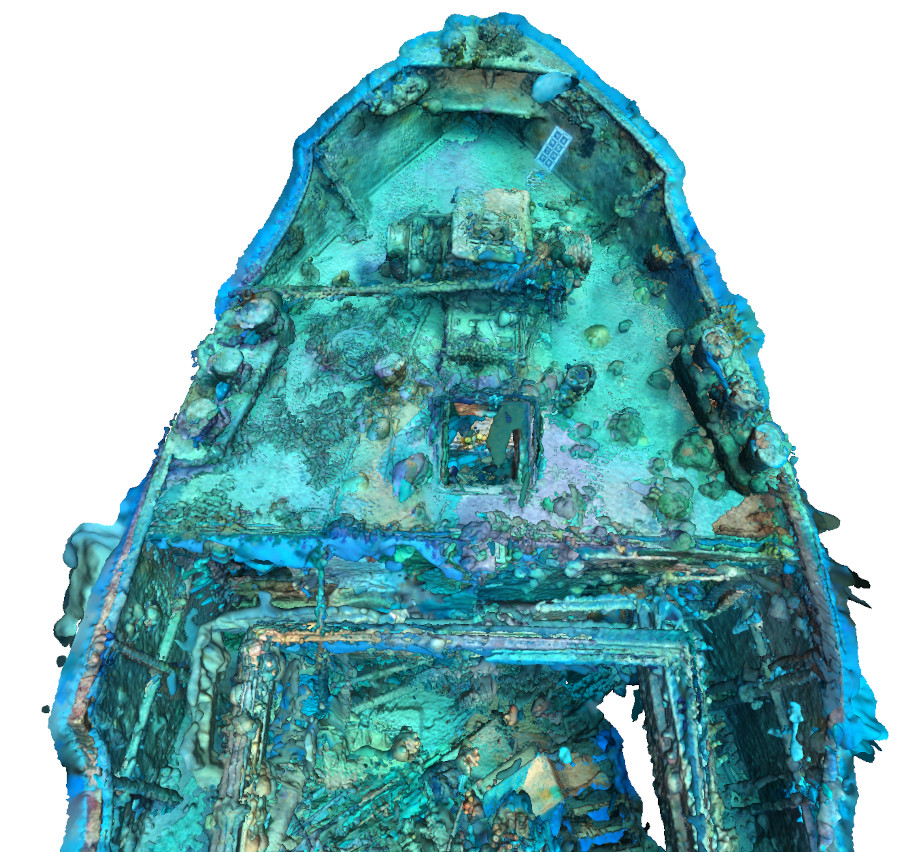}}
         \caption{} \label{DensePartsA}
   \end{subfigure}&
   \begin{subfigure}{0.31\textwidth}
          \fbox{\includegraphics[height=0.14\textheight]{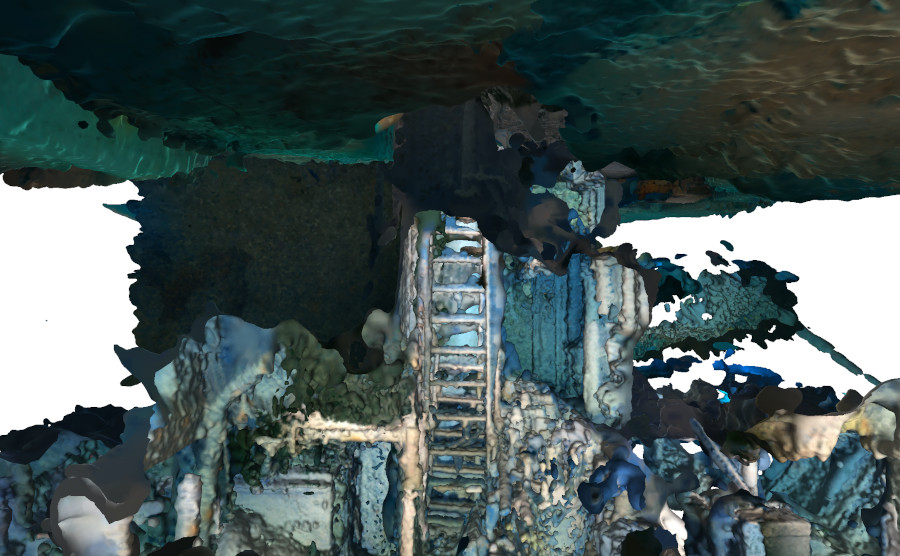}}
   \caption{} \label{DensePartsB}
   \end{subfigure}&
   \begin{subfigure}{0.31\textwidth}
          \fbox{\includegraphics[height=0.14\textheight, trim={0in, 0.0in, 0in, 0in},clip]{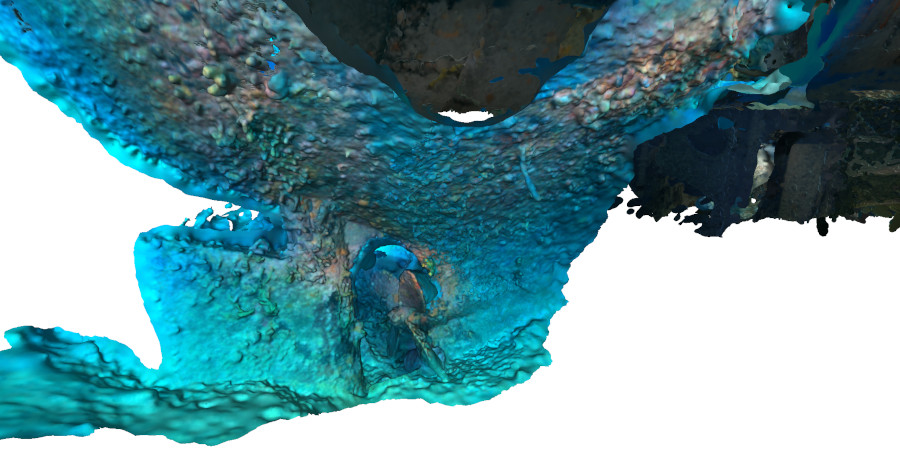 }}
   \caption{} \label{DensePartsC}
   \end{subfigure}
    \end{tabular}
    \caption{Details from the  Poisson reconstruction. (a) Top view of the bow of the wreck, visible is the crane that has fallen into the cargo hold. (b) Inside the engine room, the ladder to the corridor above is visible. (c) The rudder and the propeller at the stern of the wreck.}
    \label{fig:DenseParts}
\end{figure*}

We applied screened Poisson surface reconstruction \cite{kazhdan2013screened},  implemented by COLMAP, on the 2025 data. Views of the resulting mesh can be seen in \fig{fig:poisson}. Zoomed in details can be seen in \fig{fig:DenseParts}.  In particular, \fig{fig:DenseParts}(a) presents the Poisson reconstruction of the area shown in \fig{fig:SparseParts}(a) and (d); the  target is clearly visible at the bow. In \fig{fig:DenseParts}(b) the partial reconstruction of the interior of the engine room can be seen. Finally, \fig{fig:DenseParts}(c) presents a reconstruction of the rudder and propeller of the ship.

\section{Conclusions and future work}
The proposed framework produced accurate dense reconstructions of a shipwreck at the correct water depth. Leveraging commonly available tools enables the reconstruction out of more than \num{300000} frames by utilizing the keyframes produced by SVIn2. Even with the reduced number of images, global optimization (COLMAP) had to operate for several days in order to produce dense reconstruction. The proposed pipeline (converting video to ROS bagfile, running SVIn2, incorporating depth measurements from a dive computer, and finally feeding the produced keyframes and camera poses into COLMAP) is combined into a docker package that can be run on any platform; the code~\footnote{\url{https://github.com/AutonomousFieldRoboticsLab/Mapping_Pamir_Software}} is available as open source. 

Future work will extend the multi-sensor fusion approach~\cite{eckenhoff2021mimc,yang2024multi,zhang2020lightweight}
 to incorporate a number of asynchronous camera/IMU data streams in a common framework. As the cameras are not synchronized, the accelerometer data will be utilized to estimate the time-shift between the different cameras, while views of a calibration target will be used to estimate the extrinsics between different cameras. The possibility for a group of scientists equipped with action cameras and dive computers to accurately map a site of interest and produce models with correct scale will enable new scientific discoveries and advance the quality of environmental and infrastructure monitoring. 

\bibliographystyle{IEEEtran}
\bibliography{IEEEabrv,refs,pubs,refs_pm}

\end{document}